\pdfoutput=1
\documentclass[twoside]{amsart}
\usepackage[section]{placeins}
\usepackage[final]{microtype}

\usepackage[x11colors]{xcolor}
\usepackage{tikz}
\usetikzlibrary{decorations.pathreplacing,arrows,positioning,matrix}
\tikzset{>=stealth}
\usepackage{amsmath}
\usepackage{amssymb}
\usepackage{hyperref}
\usepackage{graphicx}

\newtheorem{thm}{Theorem}

\DeclareMathOperator{\Vol}{Vol}

\DeclareMathOperator{\argmin}{argmin}

\begin{document}


\title[CDER]{Supervised Learning of Labeled Pointcloud Differences via Cover-Tree Entropy Reduction}
\author{Abraham Smith \and
        Paul Bendich \and
        John Harer \and
        Alex Pieloch \and
        Jay Hineman}
\address{A. Smith\\
Mathematics, Statistics, and Computer Science\\
University of Wisconsin-Stout\\
Menomonie, WI, 54751, USA\\
Geometric Data Analytics, Inc. \\
Durham, NC, 27707, USA}
\email{smithabr@uwstout.edu}
\address{P. Bendich \\
Mathematics \\
Duke University \\
Durham, NC, 27708, USA \\
Geometric Data Analytics, Inc. \\
Durham, NC, 27707, USA \\
United States}
\email{bendich@math.duke.edu}
\address{J. Harer \\
Mathematics, Electrical and Computer Engineering\\
Duke University\\
Durham, NC, 27707, USA \\
Geometric Data Analytics, Inc. \\
Durham, NC, 27707, USA}
\email{harer@math.duke.edu}
\address{A. Pieloch\\
Mathematics\\
Columbia University\\
New York, NY, 10027, USA}
\email{pieloch@math.columbia.edu}
\address{J. Hineman \\
Geometric Data Analytics, Inc.\\
Durham, North Carolina, 27707, USA}
\email{jay.hineman@geomdata.com}

\date{October 24, 2016}
\begin{abstract}
We introduce a new algorithm, called CDER, for supervised machine learning
that merges the multi-scale geometric properties of Cover Trees with the
information-theoretic properties of entropy.  CDER applies to a training set of 
labeled pointclouds embedded in a common Euclidean space.  If typical pointclouds corresponding to distinct labels tend to differ
at any scale in any sub-region, CDER can identify these differences in linear time, creating a set of distributional
coordinates which act as a feature extraction mechanism for supervised learning. We describe theoretical properties and implementation details
of CDER, and illustrate its benefits on several synthetic examples.
\end{abstract}

\maketitle

\section*{Acknowledgments}

All authors partially supported by OSD contract number N00024-13-D6400 via
JHU-APL subcontract number 131753.  (Distribution Statement A - Approved for
public release; distribution is unlimited.) Bendich and Harer were also
partially supported by NSF award BIGDATA 1444791. 
Harer was also partially supported by the DARPA MoDyL program, number HR0011-16-2-0033.
Pieloch was supported by the National Science Foundation Graduate Student Fellowship Program through grant DGE 16-44869.
We thank Christopher J.
Tralie for many helpful insights about cover trees, and we thank David Porter
and Michael Williams at JHU-APL for many motivational discussions.

Our code is available at \url{https://github.com/geomdata/gda-public/} with
documentation at \url{https://geomdata.github.io/gda-public/}

\section{Overview and Assumptions}
We propose a new supervised machine-learning method for classification, where each object to be
classified is a weighted \emph{pointcloud}, and the classification task is to learn which of a finite set of labels should be applied
to the pointcloud. The method is fast, data-driven,
multi-scale, and requires no tuning.  Additionally, its details are transparently
geometric; it does not suffer from the ``black box'' interpretation
difficulties that arise in many machine-learning methods.  We call it
Cover-tree Differencing via Entropy Reduction [CDER, pronounced ``cedar''].

A high-level sketch of CDER is as follows. We construct a partial cover tree
\cite{Har-Peled} on the union of all labeled pointclouds that are given in some
training set, and we search the cover tree for convex regions that are likely
to be local minima of entropy \cite{Baez}.  For each such region, we build
distributional coordinates from the dominant labels' data.  An ordered sequence
of these distributional coordinates allows us to determine the likelihood of
each label for an unlabeled test pointcloud.

Section~\ref{sec:distcoord} explains the notion of distributional coordinates
defined on pointclouds as a front-end for supervised learning.
Section~\ref{sec:covertree} details the cover-tree algorithm and our
enhancements to it.  Section~\ref{sec:entropy} explains our simple approach to
entropy minimization in the labeled pointcloud context. Section \ref{sec:cder} gives a formal
description of CDER. Section~\ref{sec:examples} illustrates CDER with several
synthetic examples.  The remainder of this section establishes context and
notation for this problem.

Fix a Euclidean\footnote{See the Discussion section for further thoughts on this requirement.} space $\mathbb{R}^D$.  A
\emph{pointcloud} is a finite\footnote{In this document, $|X|$ indicates the
cardinality of a finite set $X$, whereas $\|v\|$ indicates a norm.} set $X = \{
x_1, x_2, \ldots, x_{|X|}\} \subset \mathbb{R}^D$.  We allow weighted
pointclouds, where each $x \in X$ has a positive weight $w(x)$; the utility of
weights is explained in Section~\ref{sec:weight}.  A \emph{cloud
collection} is a finite set of pointclouds \[\mathcal{X} = \{ X_1, X_2, \ldots,
X_{N}\},\] with $N=|\mathcal{X}|$.  Let $\underline{\mathcal{X}}=
\bigcup_{i=1}^N  X_i \subset M$ denote the union of all pointclouds in a cloud
collection, which is itself a pointcloud with weights and labels of each point
$x \in \underline{\mathcal{X}}$ inherited from the respective $X_i \owns x$.

For our supervised learning task, the training data is a cloud collection
$\mathcal{X}$, where each pointcloud is labeled from a finite set of labels,
$\Lambda$.  For the sake of discussion and visualization, we usually interpret
$\Lambda$ as a set of colors, but for mathematical purposes, $\Lambda = \{ 1,
\ldots, L\}$ where $L = |\Lambda|$.  Let $\lambda: \mathcal{X} \to \Lambda$
denote the label function, and we also use $\lambda$ to indicate a generic
label in $\Lambda$.

It is acceptable that the pointclouds $X_i$ have unequal sizes across $X_i \in
\mathcal{X}$.  It is also acceptable that the labeled sub-collections
$\mathcal{X}|_{\lambda} = \{ X_i ~:~ \lambda(X_i) = \lambda\} =
\lambda^{-1}(\lambda)$ have unequal sizes across labels $\lambda \in \Lambda$.
Pointwise weights can be assigned to compensate for these sizes, as in Section~\ref{sec:weight}.
Our only structural hypothesis is that, for each labeled sub-collection, the
pointclouds are sampled from an underlying density function---or from several
density functions as chosen by a random process---on $\mathbb{R}^D$.  For each
label $\lambda$, let $f_\lambda$ denote this density function.

\begin{figure}
\begin{center}
\includegraphics[width=0.45\columnwidth]{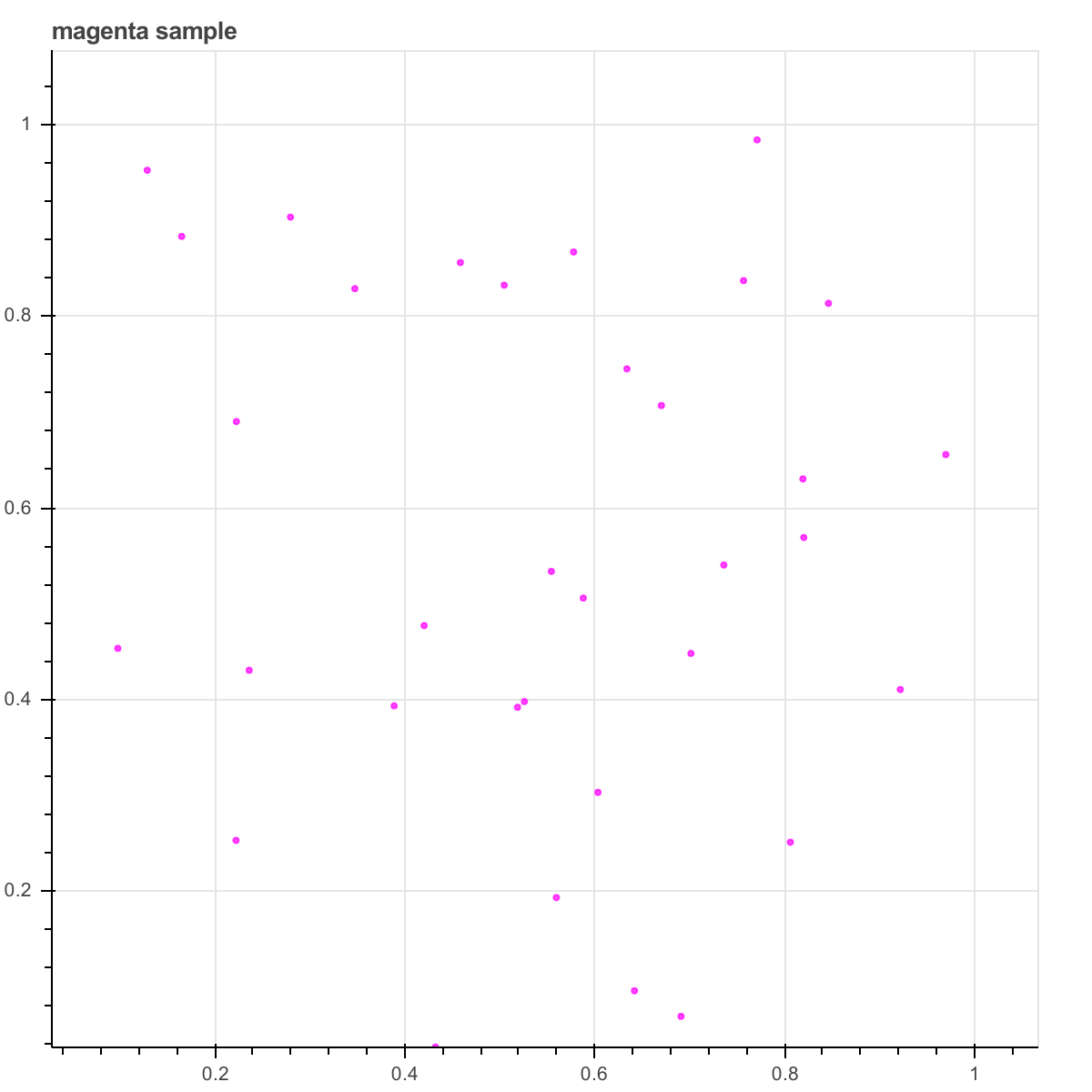}
\includegraphics[width=0.45\columnwidth]{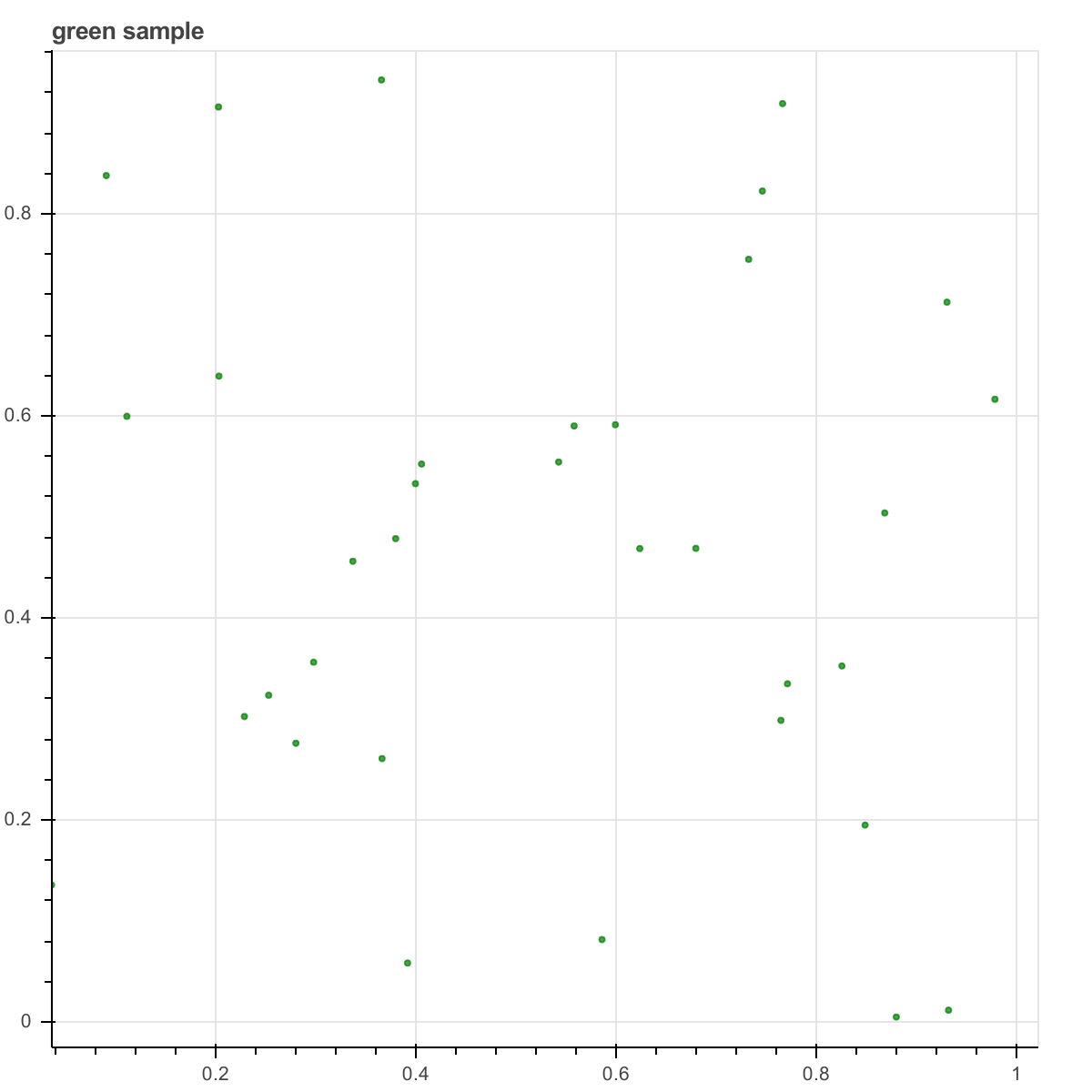}\\
\includegraphics[width=0.45\columnwidth]{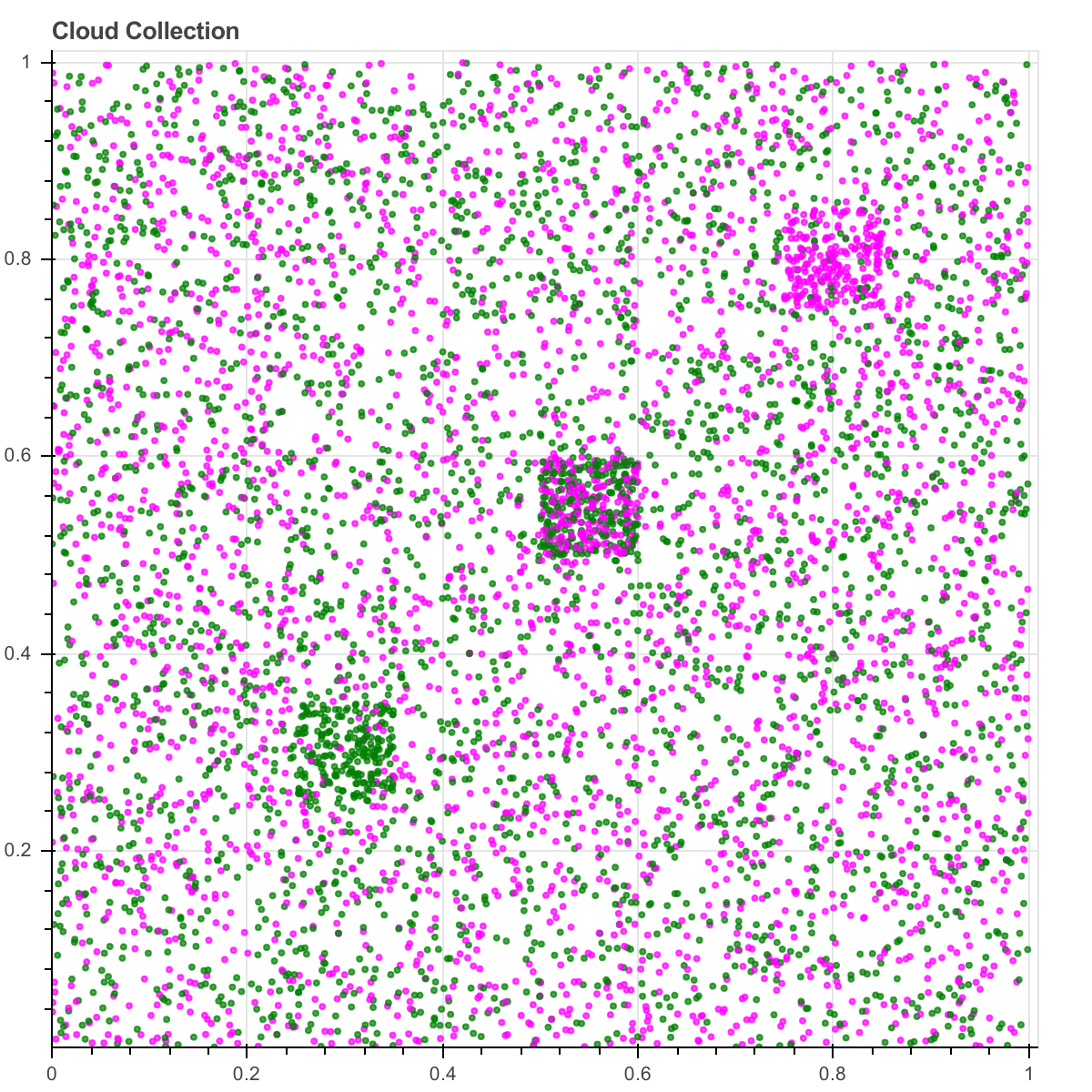}
\includegraphics[width=0.45\columnwidth]{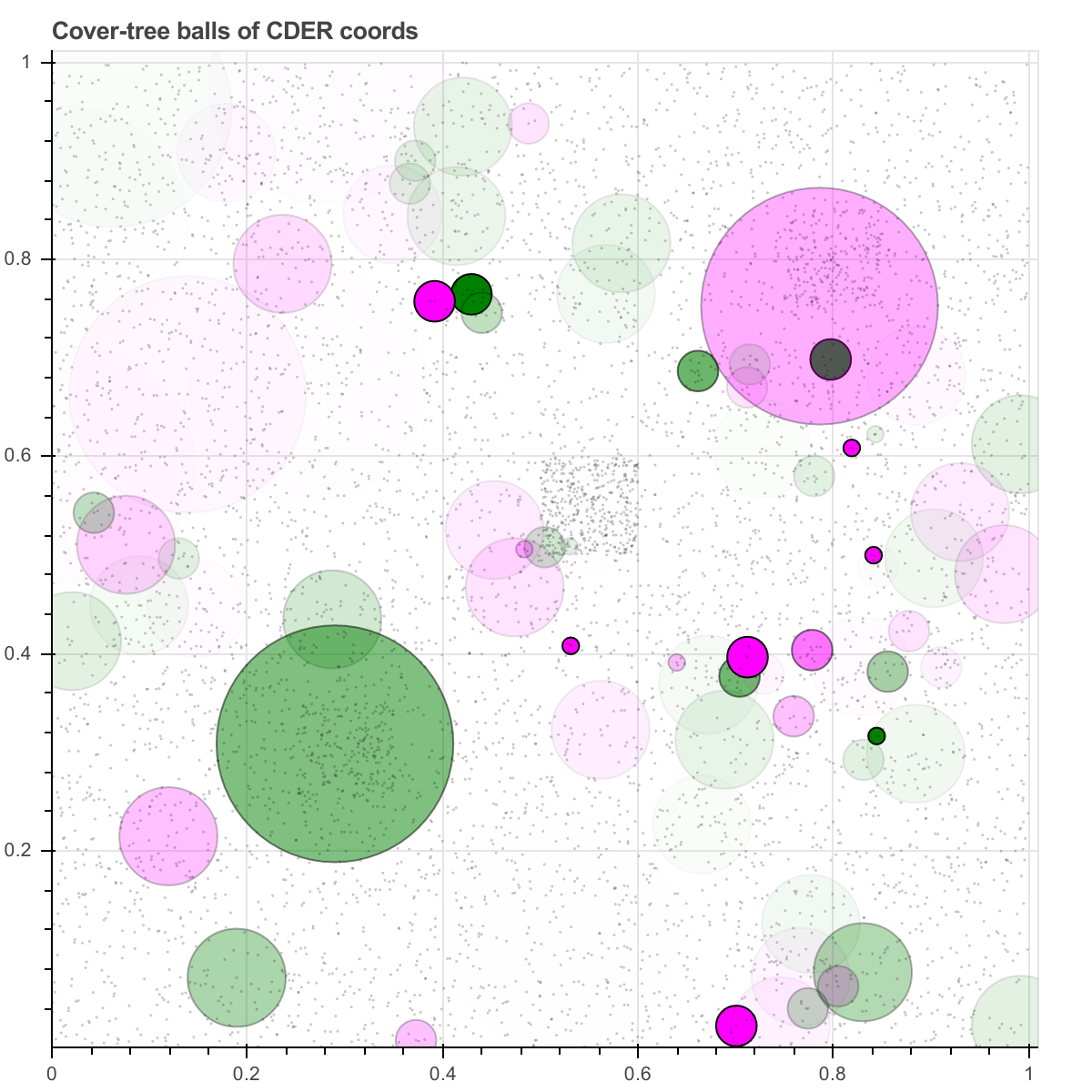}
\end{center}
\caption{Top: a single magenta pointcloud with 34 points, and a single green pointcloud with
34 points. Bottom:  the union of a
cloud collection consisting of 100 pointclouds from each
label, and CDER identifies the upper-right and lower-left
concentrations while mostly ignoring the background and the mutual
concentration in the middle.}\label{fig:blocky}
\end{figure}

We aim to efficiently find regions in $\mathbb{R}^D$ where one label has greater density than the
other labels.  That is, we seek convex regions $\Omega \subset \mathbb{R}^D$ that are
``characteristic'' of a particular label $\lambda$ because $f_\lambda$ is
unusually prominent there, while ignoring regions where the various density
functions are similar.  See Figure~\ref{fig:blocky}.

We analyze the relative density of labels over regions $\Omega$ using the
information-theoretic notion of entropy (see Section~\ref{sec:entropy}), locating subregions
of low entropy.
Because of the use of cover trees, our method emphasizes regions $\Omega$ that
are very large, very dense, or small-yet-remote (see
Section~\ref{sec:covertree}).   For each of these regions $\Omega$, we
construct a function $g$ that approximates $f_\lambda$ near $\Omega$; such a
$g$ is an example of what we call a \emph{distributional coordinate} (see
Section~\ref{sec:distcoord}).

\section{Weights}\label{sec:weight}
Typically, pointclouds are experimental samples that do not come with
pre-assigned weights, and we assign weights as follows: We have the
\emph{prior} assumption that each label $\lambda \in \Lambda$ is equally likely
among the test data.\footnote{See the discussion section for more general thoughts about this assumption.}
We give the training set total weight 1.  That is, each label
is alloted total weight $\frac{1}{L}$, where $L = |\Lambda|$.  Moreover, we make the
assumption that each sample $X_i$ is equally representative of its underlying
density function $f_{\lambda_i}$, regardless of $|X_i|$.  That is, each
pointcloud $X_i$ is alloted total weight $\frac{1}{L N_i}$, where $N_i$ is the
cardinality of ${X}|_{\lambda_i}$, the number of training pointclouds
with label $\lambda_i$.  Finally, we assume all points $x \in X_i$ have equal
weight, so $w(x) = \frac{1}{L N_i |X_i|}$.  Let $\underline{\mathcal{X}}$
denote the labeled and weighted pointcloud given by the union of all $X_i$ with
these labels and weights.

For example, suppose that the training set consists of $X_1, X_2, X_3$ where
$|X_1| = 4$, $|X_2|=10$, and $|X_3|=3$ with labels red, blue, and red,
respectively.  Suppose we want $\underline{\mathcal{X}}$ to have unit weight.
Then each $x \in X_1$ has weight $w(x) = \frac{1}{2\cdot2\cdot4} =
\frac{1}{16}$, each $x \in X_2$ has weight $w(x)= \frac{1}{2\cdot1\cdot10} =
\frac{1}{20}$, and each $x \in X_3$ has weight $w(x) = \frac{1}{2\cdot2\cdot3}=
\frac{1}{12}$.

These weights are chosen to model relative likelihoods in the following way:
If each label $\lambda$ has an underlying density function $f_\lambda$, and
each $X_i$ with label $\lambda$ is a sample from $f_\lambda$, then
$\int_\Omega f_\lambda$ should be proportional to
$\sum_{X_i \in \mathcal{X}|_{\lambda}} \sum_{x \in X_i\cap\Omega} w(x)$
for any measurable set $\Omega$.

\section{Distributional Coordinates}\label{sec:distcoord}

Given a cloud collection $\mathcal{X}$ as above, we would like to map each
pointcloud $X$ to a vector $\textbf{v}(X) \in \mathbb{R}^k$ for some fixed
dimension $k$. While this is an interesting topic in its own right, we are
mainly concerned with viewing this as a \emph{feature extraction} method: that
is,
the vectors $\textbf{v}_i = \textbf{v}(X_i) \in \mathbb{R}^k$ and corresponding
labels $\lambda_i = \lambda(X_i)$ will be used as input for any supervised
learning algorithm that requires collections of vectors in a common Euclidean
space, along with class labels.

One could imagine a supervised learning approach to labeled pointclouds that
does not perform feature extraction; for example, by first defining a
(non-Euclidean) metric between pointclouds, and then applying a
nearest-neighbor or clustering method.  However, most such metrics are unstable
to noise (such as Hausdorff distance) or are slower-than-linear to compute
(such as various Kullback-Leibler Divergence \cite{kullback1951} or the
Wasserstein metric \cite{Kerber2016wasserstein}).

This section frames the problem of feature extraction in the common language of
\emph{distributional coordinates}. We start with a definition of this intuitive
concept, and then describe the advantages and disadvantages of some standard
examples. The approach we advocate comes in the next section.

Let $g$ be an integrable function on Euclidean space.  For any
pointcloud $X = \{x_1, \ldots, x_{|X|}\},$ with pointwise weights $w_i = w(x_i)$,
define \[\int_X g = w_1 g(x_1) + \ldots + w_{|X|}g(x_{|X|}).\] We
refer to $g$ as a \emph{distributional coordinate}.  Any ordered set $(g_1,
\ldots, g_k)$ of distributional coordinates transforms $X$ into a vector
$\mathbf{v}_X = (\int_X g_1, \ldots, \int_X g_k) \in \mathbb{R}^k$.
It is computationally important to keep the dimension $k$---or at least the intrinsic
dimension of $\{\mathbf{v}_X~:~ X \in \mathcal{X}\}$---as small as possible,
as we hope to minimize the negative effect of the curse of dimensionality on the supervised learning
problem.

We use the cloud collection $\mathcal{X}$ shown on the bottom of Figure~\ref{fig:blobby} as a
notional working example in the Euclidean plane. There are two class labels,
``magenta'' and ``green.'' We imagine that each magenta pointcloud (a typical example on the top-left of the figure) contains a large
number of points sampled from a large central blob and a few points sampled
from much smaller blobs along the horizontal axis. Each green pointcloud (a typical example on the top-right of the figure) contains a
large number of points sampled from the same large central blob and a few
points sampled from smaller, different blobs along the horizontal axis.

In a typical supervised learning context, $\mathcal{X}$ would be split (say 80:20) into
training and testing sets, and we would build a set of distributional
coordinates from the labeled training set which could then infer  the correct
color labels on the testing set.

\subsection{Binning}
\begin{figure}
 \begin{center}
  \includegraphics[width=0.45\columnwidth]{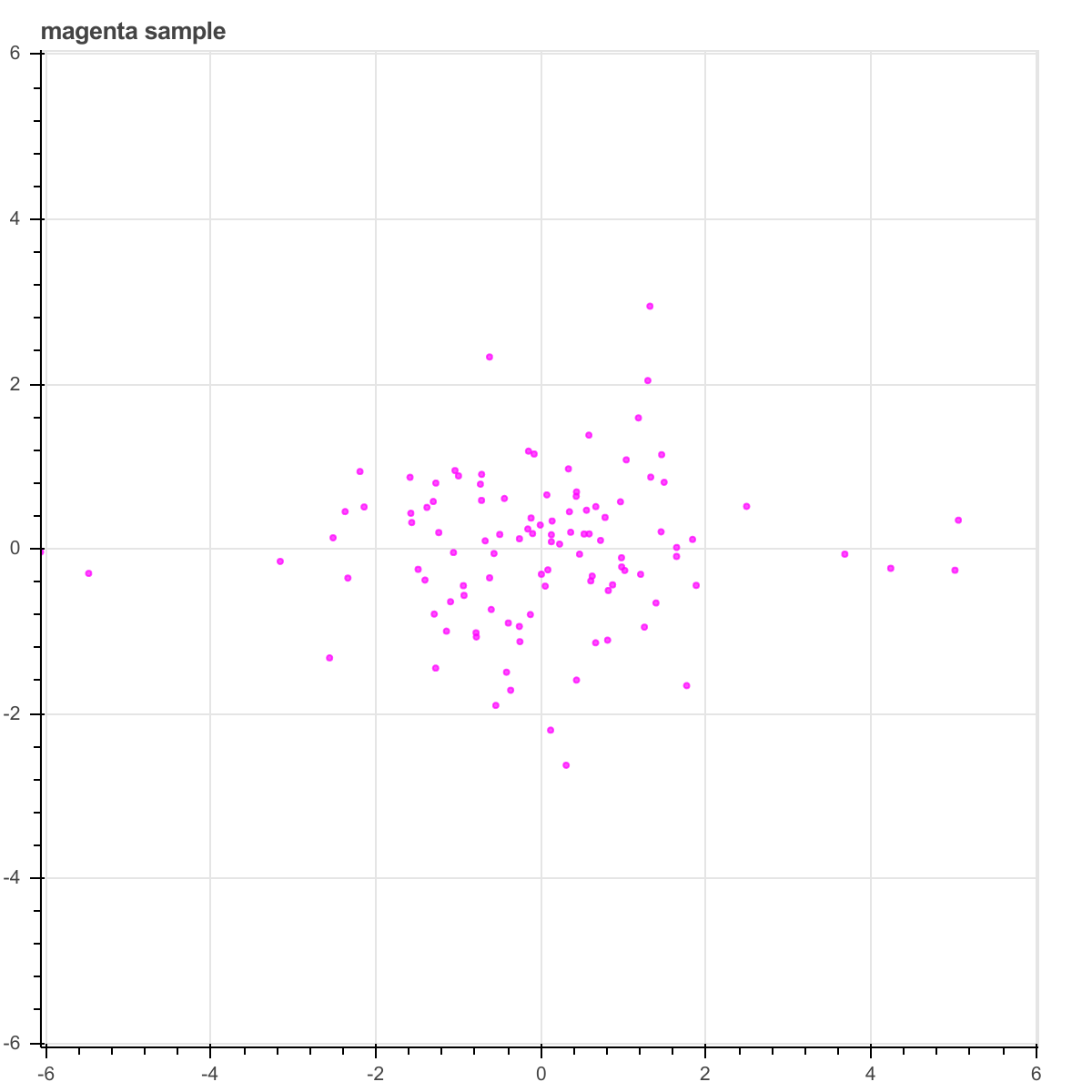}
  \includegraphics[width=0.45\columnwidth]{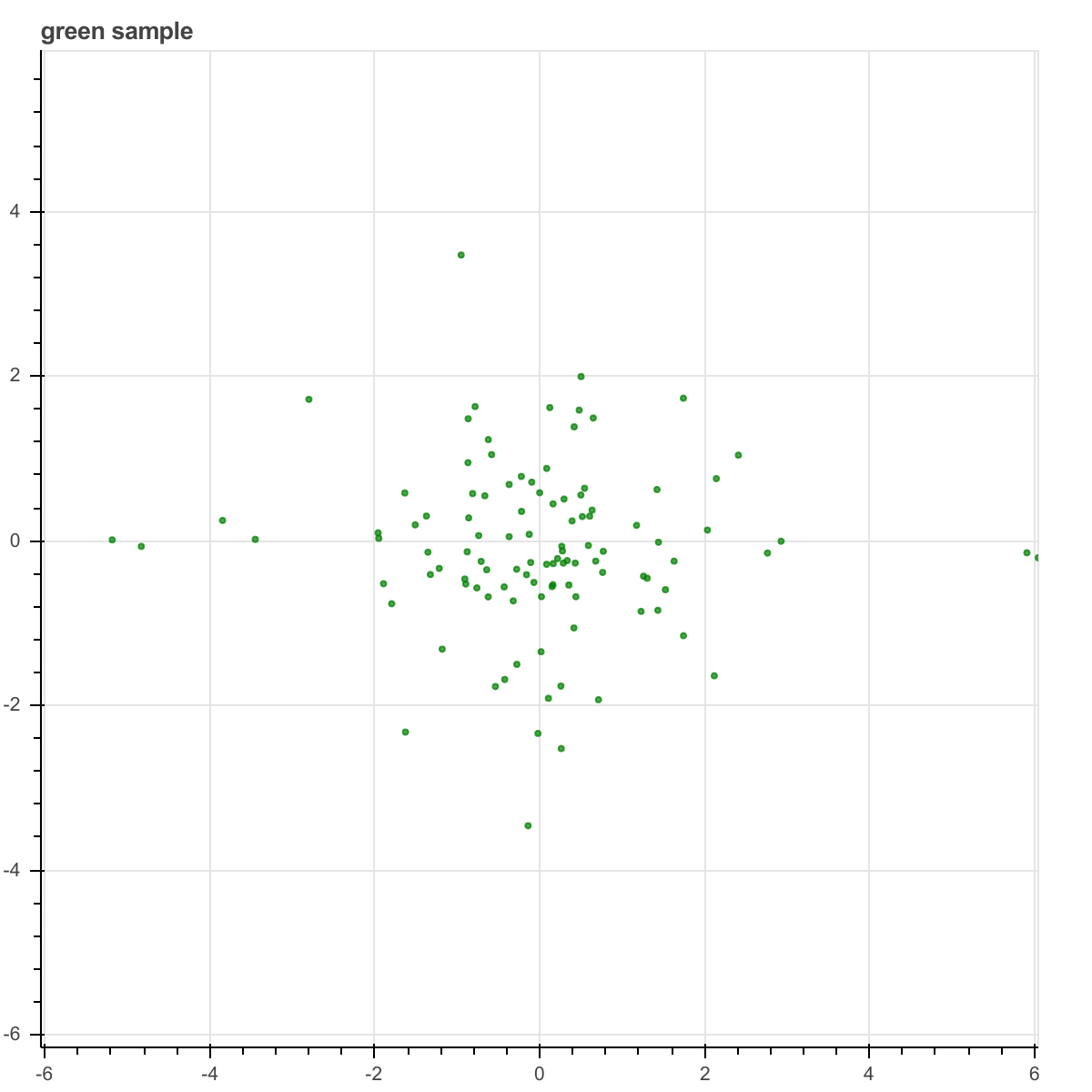}
  \includegraphics[width=0.90\columnwidth]{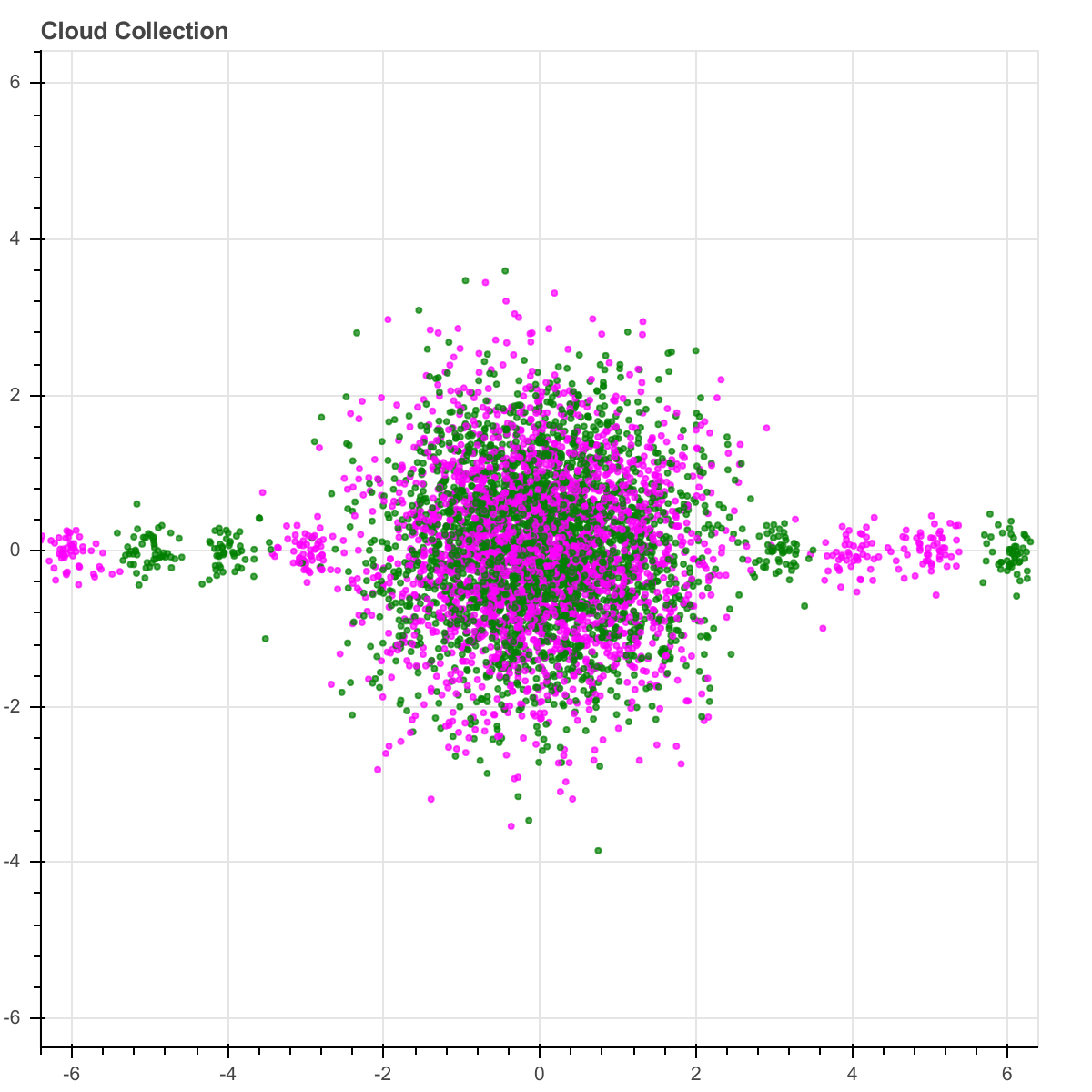}
 \end{center}
\caption{Bottom: a cloud collection of magenta and green point clouds.
Top-left: a typical magenta point cloud. Top-right a typical green point cloud.}
\label{fig:blobby}
\end{figure}

A simple approach would be to let each $g_i$ be the indicator function on some
region $R_i$; that is, $g_i$ takes the value $1$ on some region $R_i$ and $0$
elsewhere. Then $\int_X g_i$ counts the total weight of the points $X \cap R_i$. Assuming the regions $R_1, \ldots R_k$ cover the domain with overlap only on boundaries (for example, the $R_i$ could be the squares shown in Figure~\ref{fig:blobby})
this is what is commonly referred to as \emph{binning}, and it has been used (for example, \cite{Bendich2016tracking}) successfully for supervised-learning on
labeled pointclouds. On the other hand, binning is obviously unstable: consider a point $x \in X$ that lies very close to the boundary between two adjacent bins.
It also can suffer from the \emph{curse of dimensionality}: a larger number of bins than are shown in Figure \ref{fig:blobby} would be needed to distinguish the class labels.

\subsection{Gaussian Mixtures}
To address instability, one could define each $g_i$ to be a Gaussian with mean
$\mu_i$ and covariance matrix $\Sigma_i$. The obvious questions are: how many
Gaussians (modes) should be used, what are their means, and what are their
covariances?  A data-driven approach should be used to fit a Gaussian Mixture Model
(\cite{ESL01}, VI.8) to $\underline{\mathcal{X}}$.

It is apparent to the human eye that nine Gaussians (1 large green and magenta, 4
small magenta, and 4 small green) model the example in Figure~\ref{fig:blobby}.  That
is, we could ``by hand'' define a good set of distributional coordinates to establish a feature vector
$\mathbf{v} = (g_1, \ldots, g_9) : \mathcal{X} \to \mathbb{R}^9$.  Moreover,
for the purposes of classification, the central ``large'' distributional
coordinate $g_1$ is useless.  And perhaps the two innermost ``small''
Gaussians are less useful, as the remaining six distributional coordinates
are sufficient for accurate classification.  Our goal is to replicate this
observation: we construct Gaussian modes $(g_1, \ldots, g_k)$ in a
data-driven manner from $\underline{\mathcal{X}}$ that are appropriate for
use as labeled distributional coordinates for the label classification
problem.  We now proceed to the computational details.

\section{Cover Trees}\label{sec:covertree}
Cover trees were originally conceived \cite{Har-Peled} as a way to accelerate
the nearest-neighbors problem.  More recently, they have been used in
dictionary learning \cite{Allard2012} and for speeding up the analysis of the
topology of pointclouds \cite{sheehy13linear}, to pick just a few applications. This section gives basic
definitions for cover trees as well as a fast algorithm for their construction;
our account is far more detailed than usually appears in the literature and is
meant to accompany our publicly available code. Sections~\ref{sec:entropy} and~\ref{sec:cder}
use cover trees to define a set of label-driven distributional coordinates on
a cloud collection with class labels.

For any finite pointcloud\footnote{For the moment we forget that
$\underline{\mathcal{X}}$ is a union of pointclouds and simply treat it as a
single pointcloud to be studied.} $\underline{\mathcal{X}} \subset M$, consider a filtration
\[ A_0 \subset \cdots \subset A_{\ell} \subset A_{\ell+1} \subset \cdots
\subset  \underline{\mathcal{X}}.\]
An element $a \in A_{\ell}$ is called
an \emph{adult}\footnote{To reduce confusion (hopefully) by maintaining a
consistent metaphor, we diverge from
prior works' notation in two places.  Our \emph{adult} is called a
\emph{center} elsewhere.  Our \emph{guardian} is usually called a
\emph{parent}.  We avoid the word parent because there are many distinct tree
relations embedded in this algorithm.} at level $\ell$.  The set $A_{\ell} \setminus A_{\ell-1}$ is the
\emph{cohort} at $\ell$.

\begin{figure}
\begin{center}
\includegraphics[width=0.45\columnwidth]{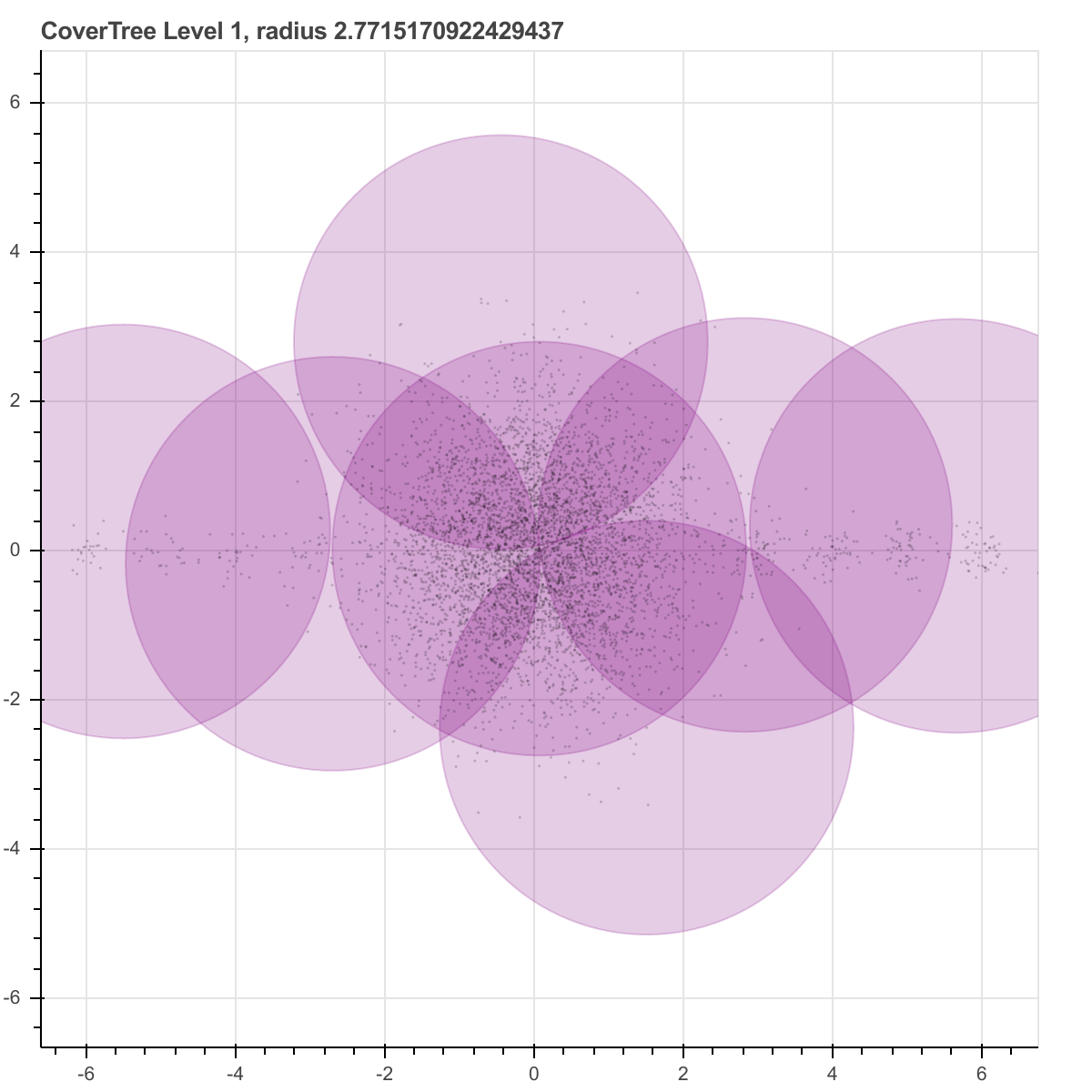}
\includegraphics[width=0.45\columnwidth]{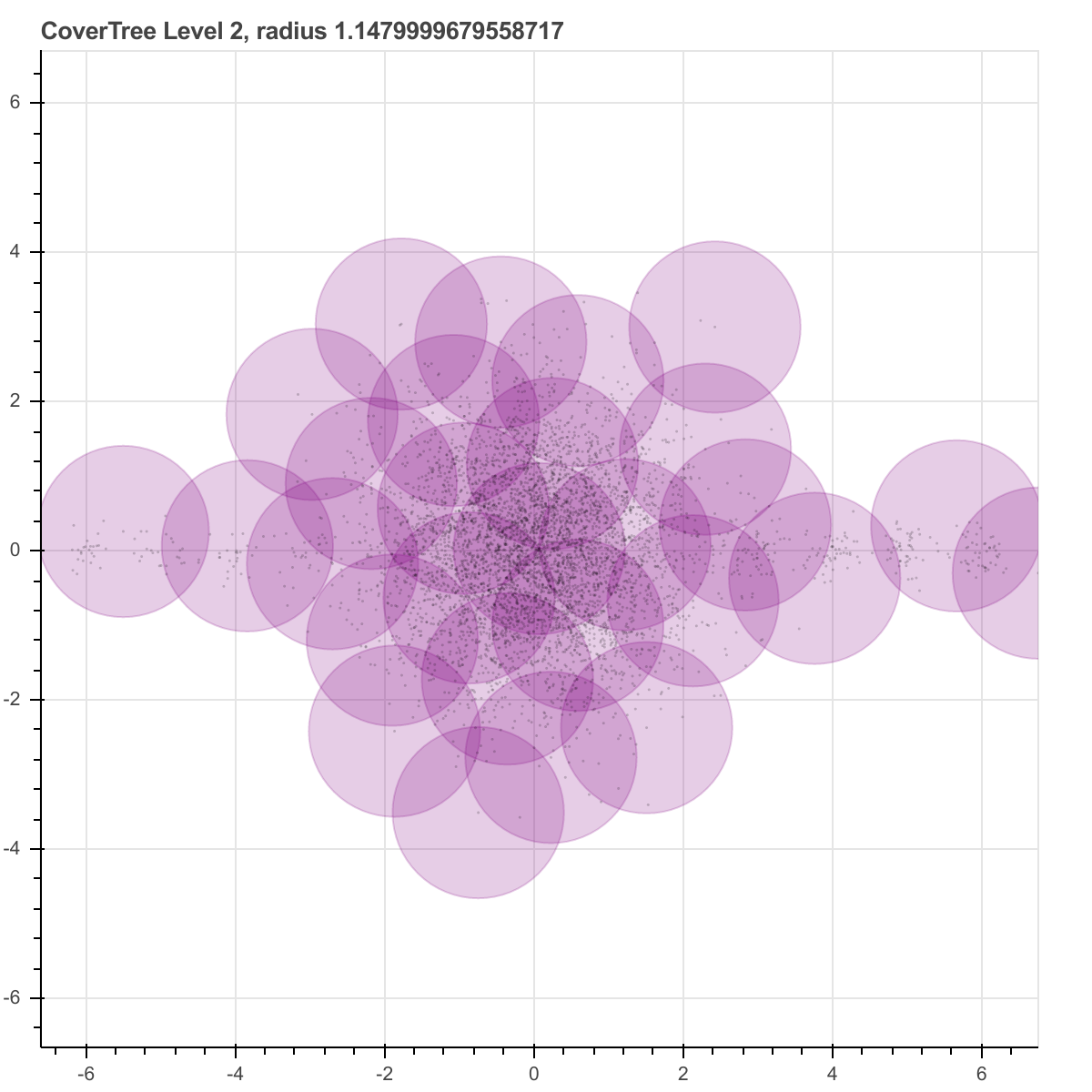}
\includegraphics[width=0.45\columnwidth]{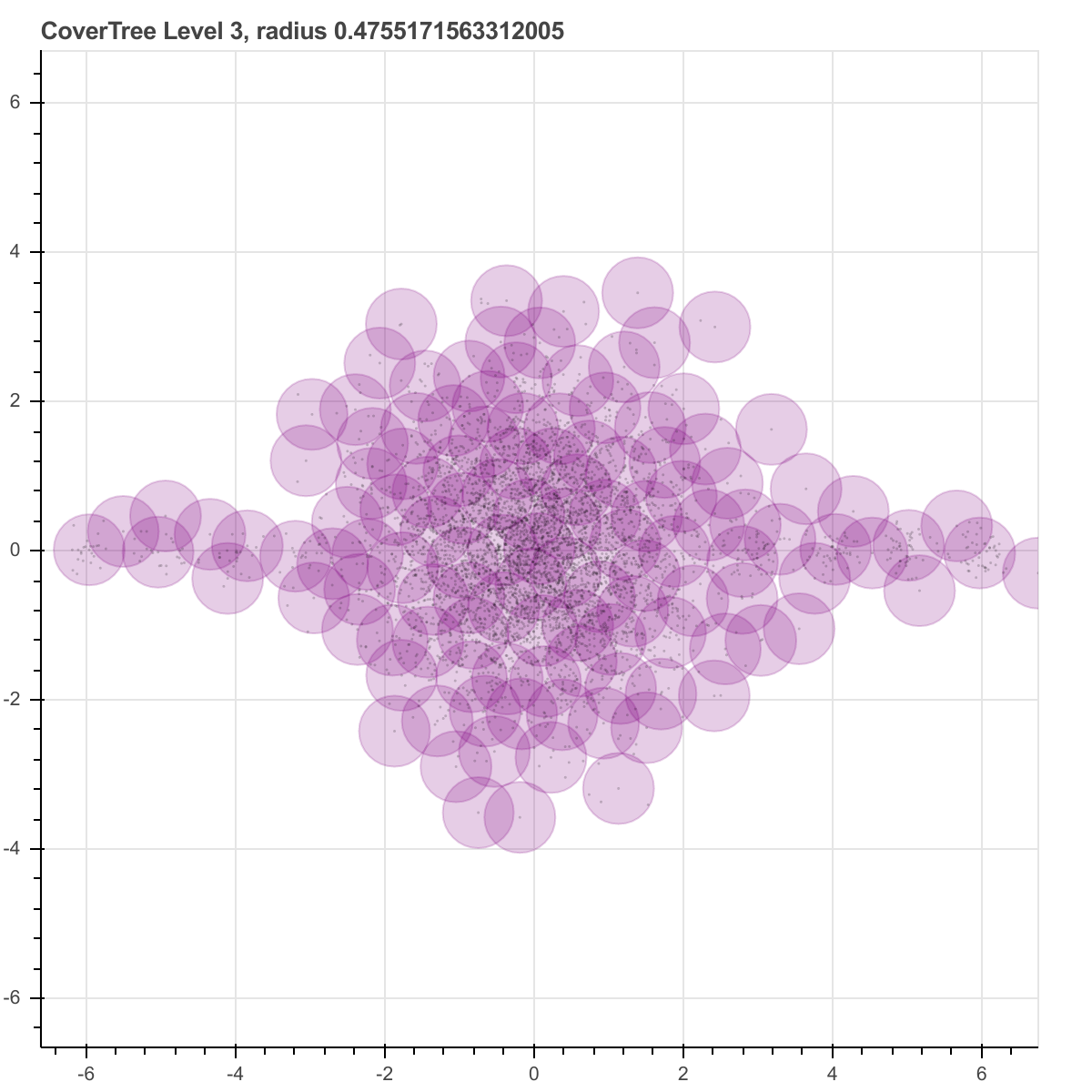}
\includegraphics[width=0.45\columnwidth]{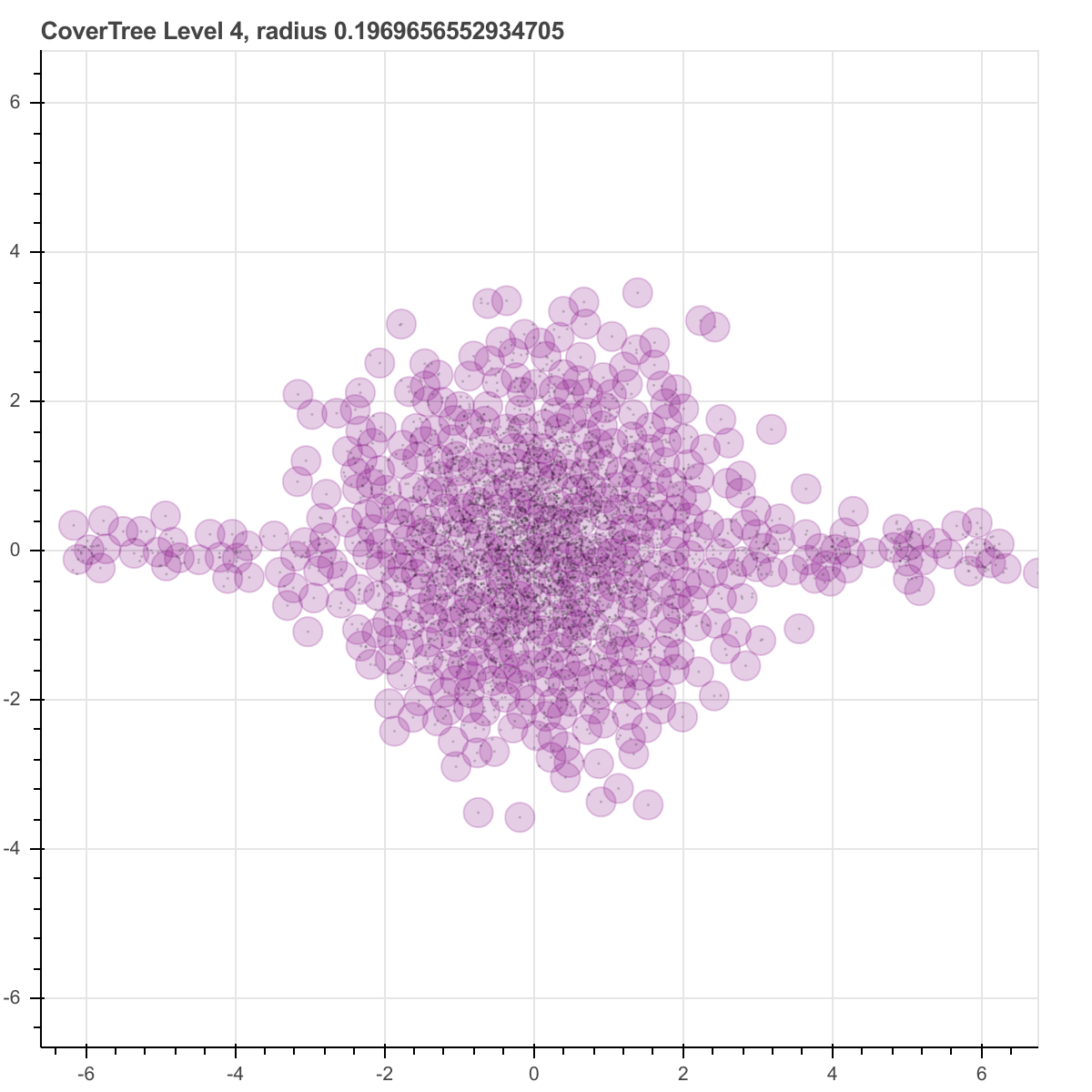}
\includegraphics[width=0.45\columnwidth]{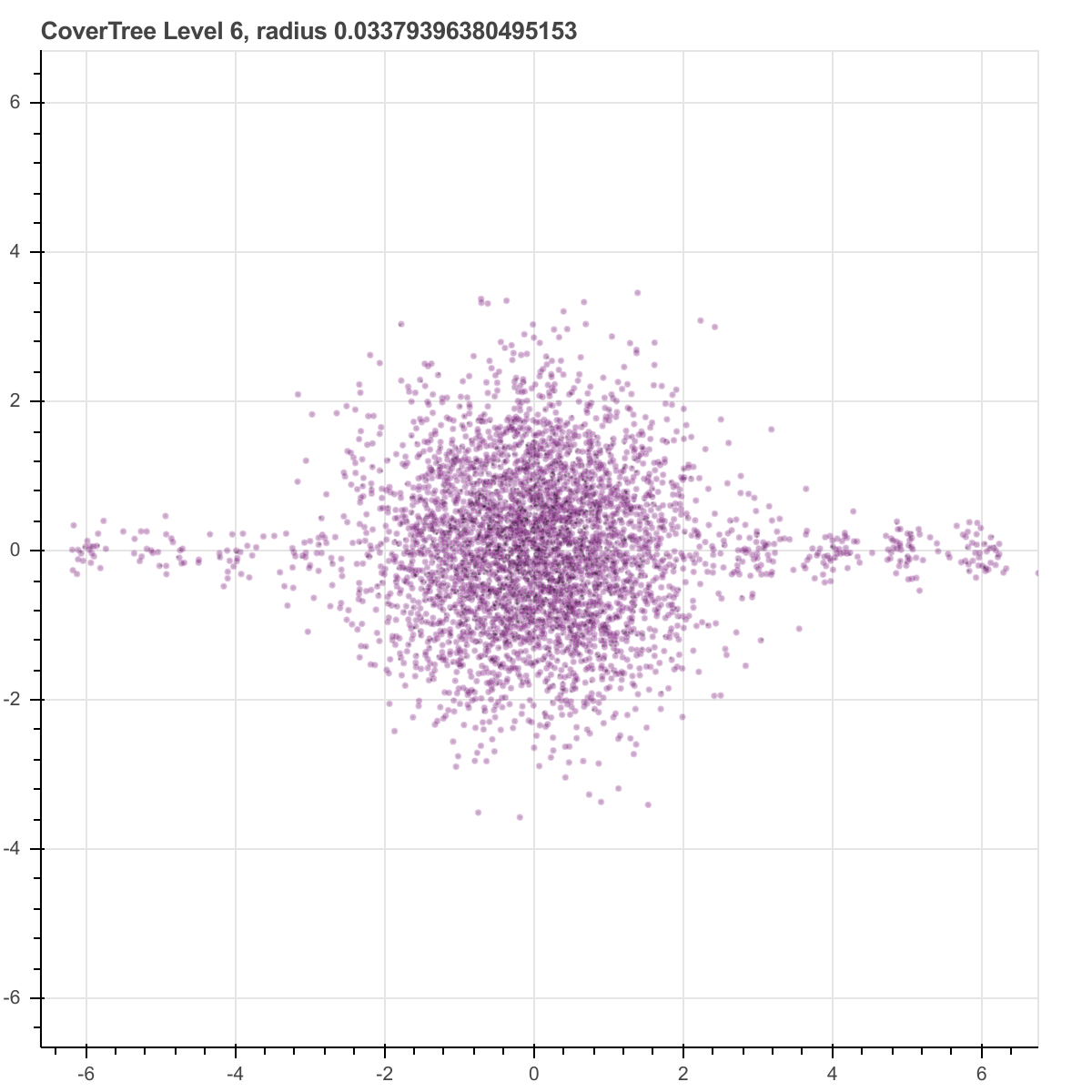}
\includegraphics[width=0.45\columnwidth]{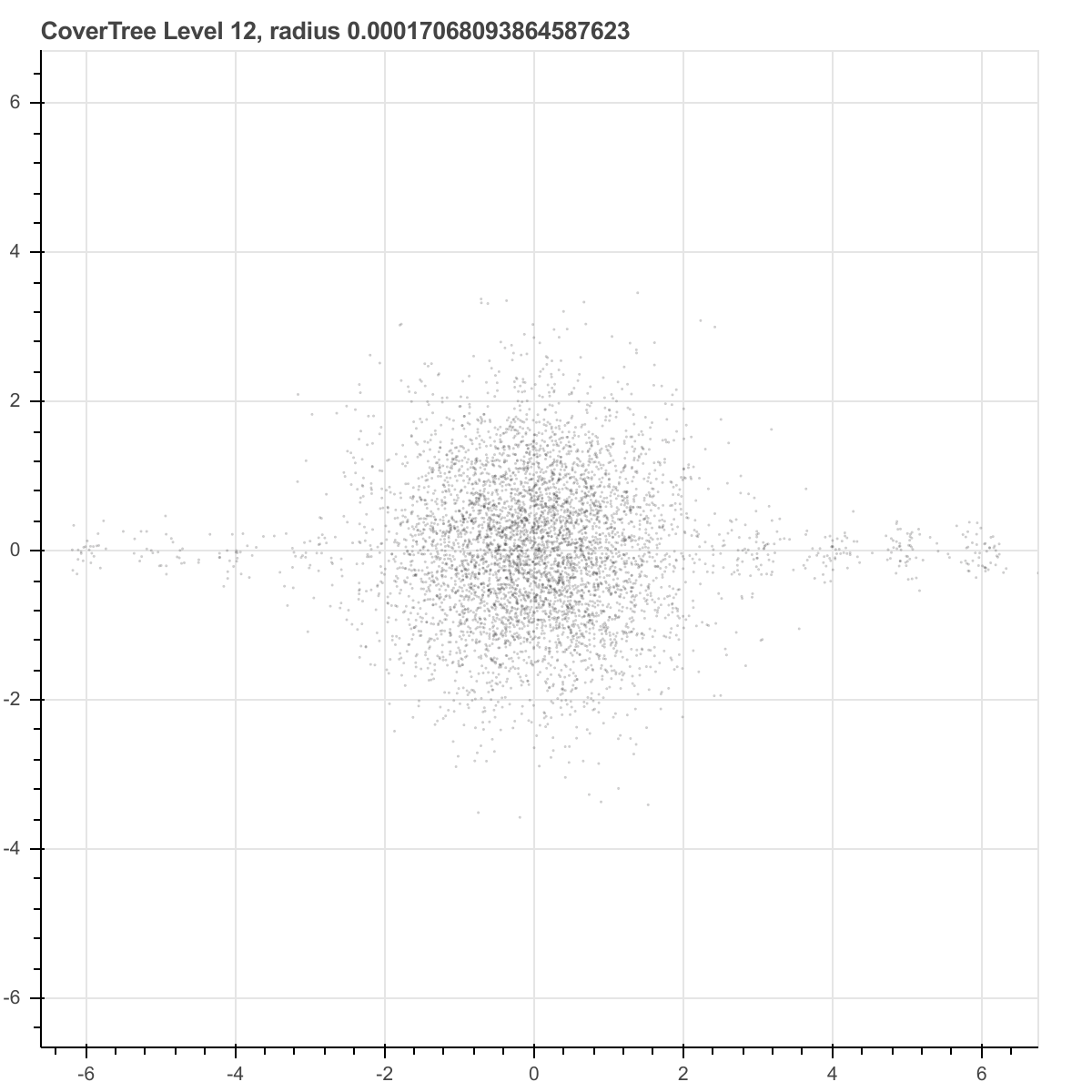}
\end{center}
\caption{Levels 1, 2, 3, 4, 6, and 12 of a cover tree for the dataset
$\underline{\mathcal{X}}$ seen in Figure~\ref{fig:blobby}.}\label{fig:covertree}
\end{figure}

A cover tree builds a filtration by covering $\underline{\mathcal{X}}$ with balls of smaller and
smaller radius centered at the points in $A_{\ell}$.
See Figure~\ref{fig:covertree}.

Specifically, a \emph{cover tree} is a filtration of $\underline{\mathcal{X}}$ with the following
additional properties:
\begin{enumerate}
\item $|A_0|=1$.  Let $A_0 = \{a_0\}$;
\item There is a radius $r_0$ such that $\underline{\mathcal{X}} \subset B(a_0, r_0)$;
\item There is a real number $0< \theta < 1$, called the \emph{shrinkage
ratio} such that, for every $\ell$,
$\underline{\mathcal{X}} \subset \cup_{a_i \in A_\ell} B(a_i, r_\ell)$ where $r_\ell =
\theta^\ell r_0$.
\item For each $\ell$, if $a_i, a_j \in A_\ell$, then $d(a_i, a_j) > r_\ell$.
\item For each $\ell$, each point $x \in \underline{\mathcal{X}}$ is assigned to a
\emph{guardian} $a_i \in A_\ell$ such that $x$ lies in the ball $B(a_i, r_\ell)$; we say $x$ is
a \emph{child} of $a_i$.  Each $a_i \in A_\ell$ is its own guardian and its own
child.
\item There is a tree structure on the (level, adult) pairs of the filtration
$(\ell, a_i)$, where there is a tree relation $(\ell, a_i) \to (\ell+1, a_k)$ if
$a_k$ was a child of $a_i$ at level $\ell$.  We say $a_k$ is a
\emph{successor} of $a_i$, and $a_i$ is a \emph{predecessor} of $a_k$.
Note that $(\ell, a_i) \to (\ell+1, a_i)$ for all $a_i \in A_\ell$.
\end{enumerate}

\subsection{Vocabulary}
Cover trees can be generated using a fast algorithm relying on the notion of
\emph{friends}, first introduced in \cite{Har-Peled}.  The rough algorithm is
familiar to many practitioners, but is not widely published.  We enhance this
algorithm by giving each point a \emph{weight} and a \emph{label}.   We
describe our method here because the details of the construction are crucial to
the supervised-learning task of this article.

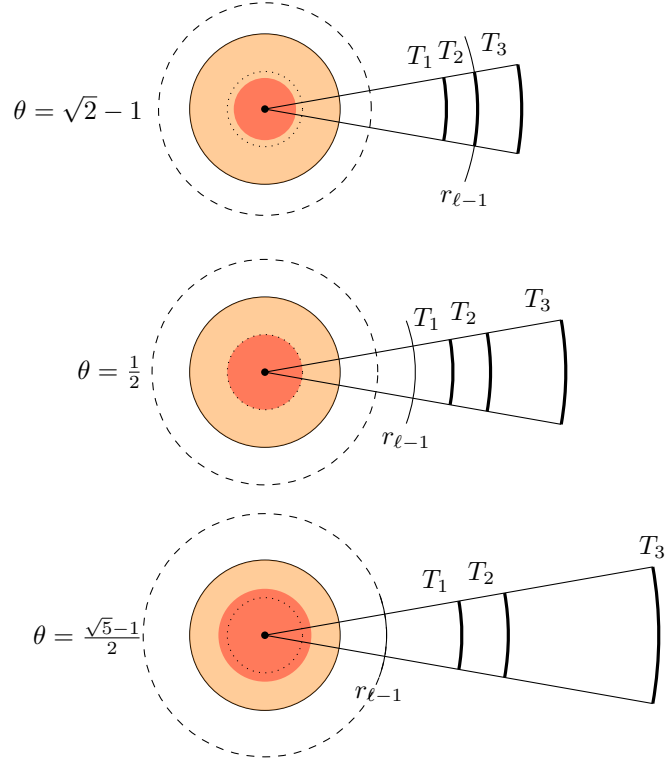
\begin{figure}
\begin{center}
\begin{tikzpicture}[scale=0.1]
\begin{scope}[shift={(0,0)}] \node[left] at (-15,0) {$\theta=\sqrt{2}-1$};
\draw (0,0) circle (10cm);
\fill[opacity=0.4, color=orange] (0,0) circle (10cm);
\fill[opacity=0.4, color=red] (0,0) circle (04.14cm);
\draw[dashed] (0,0) circle (14.14cm); 
\node [above left] at ({24.14*cos(10)}, {24.14*sin(10)}) {$T_1$};
\draw[very thick] ({24.14*cos(-10)}, {24.14*sin(-10)}) arc (-10:10:24.14);
\node [above left] at ({28.28*cos(10)}, {28.28*sin(10)}) {$T_2$};
\draw[very thick] ({28.28*cos(-10)}, {28.28*sin(-10)}) arc (-10:10:28.28);
\node [above left] at ({34.14*cos(10)}, {34.14*sin(10)}) {$T_3$};
\draw[very thick] ({34.14*cos(-10)}, {34.14*sin(-10)}) arc (-10:10:34.14);
\draw ({34.14*cos(-10)}, {34.14*sin(-10)}) --  (0,0) --  ({34.14*cos(10)},{34.14*sin(10)});
\node [below] at ({28.28*cos(-20)}, {28.28*sin(-20)}) {$r_{\ell-1}$};
\draw ({28.28*cos(-20)}, {28.28*sin(-20)}) arc (-20:20:28.28);
\draw[dotted] (0,0) circle (5cm);
\fill (0,0) circle (0.5cm);
\end{scope}
\begin{scope}[shift={(0,-35)}] \node[left] at (-15,0) {$\theta=\frac{1}{2}$};
\draw (0,0) circle (10cm);
\fill[opacity=0.4, color=orange] (0,0) circle (10cm);s
\fill[opacity=0.4, color=red] (0,0) circle (05.00cm);
\draw[dashed] (0,0) circle (15.00cm); 
\node [above left] at ({25.00*cos(10)}, {25.00*sin(10)}) {$T_1$};
\draw[very thick] ({25.00*cos(-10)}, {25.00*sin(-10)}) arc (-10:10:25.00);
\node [above left] at ({30.00*cos(10)}, {30.00*sin(10)}) {$T_2$};
\draw[very thick] ({30.00*cos(-10)}, {30.00*sin(-10)}) arc (-10:10:30.00);
\node [above left] at ({40.00*cos(10)}, {40.00*sin(10)}) {$T_3$};
\draw[very thick] ({40.00*cos(-10)}, {40.00*sin(-10)}) arc (-10:10:40.00);
\draw ({40.00*cos(-10)}, {40.00*sin(-10)}) --  (0,0) --  ({40.00*cos(10)},{40.00*sin(10)});
\node [below] at ({20.00*cos(-20)}, {20.00*sin(-20)}) {$r_{\ell-1}$};
\draw ({20.00*cos(-20)}, {20.00*sin(-20)}) arc (-20:20:20.00);
\draw[dotted] (0,0) circle (5cm);
\fill (0,0) circle (0.5cm);
\end{scope}
\begin{scope}[shift={(0,-70)}] \node[left] at (-15,0) {$\theta=\frac{\sqrt{5}-1}{2}$};
\draw (0,0) circle (10cm);
\fill[opacity=0.4, color=orange] (0,0) circle (10cm);
\fill[opacity=0.4, color=red] (0,0) circle (06.18cm);
\draw[dashed] (0,0) circle (16.18cm); 
\node [above left] at ({26.18*cos(10)}, {26.18*sin(10)}) {$T_1$};
\draw[very thick] ({26.18*cos(-10)}, {26.18*sin(-10)}) arc (-10:10:26.18);
\node [above left] at ({32.36*cos(10)}, {32.36*sin(10)}) {$T_2$};
\draw[very thick] ({32.36*cos(-10)}, {32.36*sin(-10)}) arc (-10:10:32.36);
\node [above] at ({52.36*cos(10)}, {52.36*sin(10)}) {$T_3$};
\draw[very thick] ({52.36*cos(-10)}, {52.36*sin(-10)}) arc (-10:10:52.36);
\draw ({52.36*cos(-10)}, {52.36*sin(-10)}) --  (0,0) -- ({52.36*cos(10)},{52.36*sin(10)});
\node [below] at ({16.18*cos(-20)}, {16.18*sin(-20)}) {$r_{\ell-1}$};
\draw ({16.18*cos(-20)}, {16.18*sin(-20)}) arc (-20:20:16.18);
\draw[dotted] (0,0) circle (5cm);
\fill (0,0) circle (0.5cm);
\end{scope}
\end{tikzpicture}
\end{center}
\caption{The relevant radii for the special ratios $\theta = \sqrt{2}-1$,
$\theta=1/2$, and $\theta = (\sqrt{5}-1)/2$.  The orange ball has radius
$r_\ell$.  The red ball has radius $r_{\ell+1}$.  The small dotted sphere separates youngins from teens.  The larger
dashed sphere contains the elders.}\label{fig:radii}
\end{figure}

Extending the maturation/reproduction metaphor of \emph{adults},
\emph{children}, and \emph{guardians} above, a child $x$ with guardian $a_i$ at
level $\ell$ is called a \emph{teen} if $\frac12 r_\ell < d(a_i, x)$, and it is
called a \emph{youngin} if $d(a_i, x) \leq \frac12 r_\ell$.  See
Figure~\ref{fig:radii}.
When changing from level $\ell$ to level $\ell+1$, the radius of each ball shrinks to
$r_{\ell+1} = \theta r_\ell$.  Children farther than $r_{\ell+1}$ from their guardians
become \emph{orphans}.  We must decide whether these orphans should be
\emph{adopted} by other adults at level $\ell+1$, or if the orphans should be
\emph{emancipated} as new adults at level $\ell+1$.
That is, the newly emancipated adults at level $\ell+1$ comprise the
\emph{cohort} at level $\ell+1$.


Our algorithm uses several notions of friends, set by the following bounds
at level $\ell$:
\begin{equation}
\begin{split}
T_1(\ell) &= (2 + \theta)r_\ell,\\
T_2(\ell) &= (2 + 2\theta)r_\ell,\\
T_3(\ell) &= 2( 1 + \theta + \theta^2 + \cdots ) r_\ell =
\frac{2}{1-\theta}r_\ell.
\end{split}
\end{equation}
For each $k=1,2,3$, we say that adults $a_i$ and $a_j$ at level $\ell$ are
type-$k$ \emph{friends} if $d (a_i, a_j) \leq T_k(\ell)$.
It is easy to verify the following recursion relations:
\begin{equation}
\begin{cases}
T_1(\ell) \leq T_2(\ell) \leq T_3(\ell),\\
T_1(\ell+1) \leq T_1(\ell),\\
T_2(\ell+1) \leq T_2(\ell),\\
T_3(\ell+1) \leq T_3(\ell).\\
\end{cases}
\end{equation}

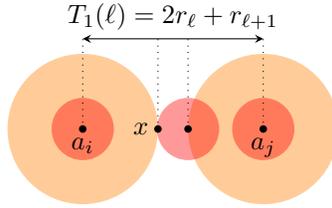
\begin{figure}
\begin{center}
\begin{tikzpicture}[scale=0.1]
\fill[opacity=0.4, color=orange] (0,0) circle (10.00cm);
\fill[opacity=0.4, color=red] (0,0) circle (04.14cm);
\fill[opacity=0.4, color=orange] (24,0) circle (10.00cm);
\fill[opacity=0.4, color=red] (24,0) circle (04.14cm);
\fill[opacity=0.4, color=red] (14,0) circle (04.14cm);

\fill (0,0) circle (0.5cm);
\node[below] at (0,0) {$a_i$};

\fill (24,0) circle (0.5cm);
\node[below] at (24,0) {$a_j$};

\fill (14,0) circle (0.5cm);

\fill (10,0) circle (0.5cm);
\node[left] at (10,0) {$x$};

\draw[<->] (0,12) -- (24,12);
\draw[dotted] (0,0) -- (0,12);
\draw[dotted] (24,0) -- (24,12);
\draw[dotted] (14,0) -- (14,12);
\draw[dotted] (10,0) -- (10,12);
\node[above] at (12,12) {$T_1(\ell) = 2r_\ell + r_{\ell+1}$};
\end{tikzpicture}
\end{center}
\caption{An orphan $x$ of $a_i$ could potentially be adopted by a successor of
$a_j$ if $d(a_i, d_j) \leq T_1$.}\label{fig:T1}
\end{figure}

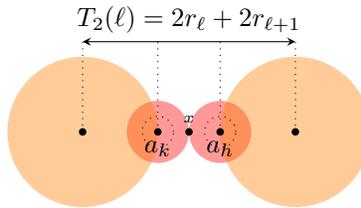
\begin{figure}
\begin{center}
\begin{tikzpicture}[scale=0.1]
\fill[opacity=0.4, color=red] (0,0) circle (04.14cm);
\fill[opacity=0.4, color=red] (08.28,0) circle (04.14cm);
\fill[opacity=0.4, color=orange] (-10,0) circle (10.00cm);
\fill[opacity=0.4, color=orange] (18.28,0) circle (10.00cm);
\fill (0,0) circle (0.5cm);
\node[below] at (0,0) {$a_k$};
\draw[dotted] (0,0) circle (02.07cm);
\fill (04.14,0) circle (0.5cm);
\node[above] at (04.14,0) {\tiny $x$};
\fill (08.28,0) circle (0.5cm);
\node[below] at (08.28,0) {$a_h$};
\draw[dotted] (08.28,0) circle (02.07cm);
\fill (-10,0) circle (0.5cm);
\fill (18.28,0) circle (0.5cm);
\draw[<->] (-10,12) -- (18.28,12);
\draw[dotted] (-10,0) -- (-10,12);
\draw[dotted] ( 0,0) -- (0,12);
\draw[dotted] (08.28,0) -- (08.28,12);
\draw[dotted] (18.28,0) -- (18.28,12);
\node[above] at (4.14,12) {$T_2(\ell) = 2r_\ell + 2r_{\ell+1}$};
\end{tikzpicture}
\end{center}
\caption{A teen $x$ of $a_k$ at level $\ell+1$ could be exchanged to some $a_h$,
where the distance between the level-$\ell$ predecessors of $a_k$ and $a_h$ is
at most $T_2(\ell)$.}\label{fig:T2}
\end{figure}

\begin{figure}
\begin{center}
\begin{tikzpicture}[scale=0.1]
\fill[opacity=0.4, color=orange] (0,0) circle (10cm);
\fill[opacity=0.4, color=orange] (34.14,0) circle (10cm);
\fill[opacity=0.4, color=red] (10.00,0) circle (04.14cm);
\fill[opacity=0.4, color=red] (24.14,0) circle (04.14cm);
\draw[very thick] ({34.14*cos(-10)}, {34.14*sin(-10)}) arc (-10:10:34.14);
\draw ({34.14*cos(-10)}, {34.14*sin(-10)}) --  (0,0) --  ({34.14*cos(10)},{34.14*sin(10)});
\draw[very thick] ({10+14.14*cos(-10)}, {14.14*sin(-10)}) arc (-10:10:14.14);
\draw ({10 + 14.14*cos(-10)}, {14.14*sin(-10)}) --  (10,0) -- ({10 + 14.14*cos(10)},{14.14*sin(10)});
\fill (0,0) circle (0.5cm);
\fill (10,0) circle (0.5cm);
\fill (24.14,0) circle (0.5cm);
\fill (34.14,0) circle (0.5cm);
\draw[<->] (0,12) -- (34.14,12);
\draw[dotted] ( 0,0) -- (0,12);
\draw[dotted] ( 10,0) -- (10,12);
\draw[dotted] (24.14,0) -- (24.14,12);
\draw[dotted] (34.14,0) -- (34.14,12);
\node[above] at (17.07,12) {$T_3(\ell) = 2r_\ell + T_3(\ell+1)$};
\end{tikzpicture}
\end{center}
\caption{If two adults are type-3 friends at level $\ell+1$, then their
predecessors are type-3 friends at level $\ell$.}\label{fig:T3}
\end{figure}
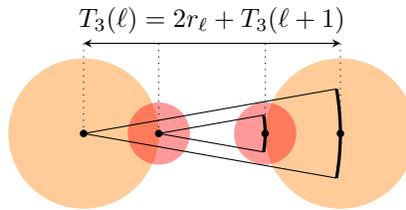

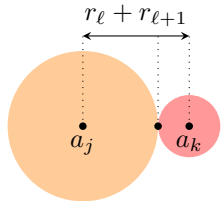
\begin{figure}
\begin{center}
\begin{tikzpicture}[scale=0.1]
\fill[opacity=0.4, color=orange] (0,0) circle (10cm);
\fill[opacity=0.4, color=red] (14.14,0) circle (04.14cm);
\fill (0,0) circle (0.5cm);
\fill (10,0) circle (0.5cm);
\fill (14.14,0) circle (0.5cm);
\node[below] at (0,0) {$a_j$};
\node[below] at (14.14,0) {$a_k$};
\draw[<->] (0,12) -- (14.14,12);
\draw[dotted] ( 0,0) -- (0,12);
\draw[dotted] ( 10,0) -- (10,12);
\draw[dotted] (14.14,0) -- (14.14,12);
\node[above] at (7.07,12) {$r_\ell + r_{\ell+1}$};
\end{tikzpicture}
\end{center}
\caption{$a_j$ is an elder of $a_k$ if a level-$(\ell+1)$ child of $a_k$ could potentially have
been a level-$\ell$ child of $a_j$.}\label{fig:elder}
\end{figure}

Here are the essential facts about friends, which are immediate consequences of
the triangle inequality:
\begin{thm}
\begin{enumerate}
\item[T1] If $x$ is a child of $a_i$ at level $\ell$ that is orphaned at level
$\ell+1$, then $x$ can be adopted only by either the type-1 friends of $a_i$ at
level $\ell$ or the newly-emancipated children at level $\ell+1$ of those type-1
friends.  See Figure~\ref{fig:T1}.

\item[T2] If we choose to re-assign children to the nearest adult, then teen $x$
with guardian $a_k$ can be adopted by adult $a_h$ at level $\ell+1$, where the
level-$\ell$ predecessors of $a_k$ and $a_h$ are type-2 friends.
See Figure~\ref{fig:T2}.

\item[T3] If $a_k$ and $a_h$ are type-3 friends at level $\ell+1$, then their
predecessors $a_i$ and $a_j$ are type-3 friends at level $\ell$.
See Figure~\ref{fig:T3}.
\end{enumerate}
\label{thm:friends}
\end{thm}

\begin{thm}
Suppose $a_k$ is an adult at level $\ell{+}1$ with predecessor $a_i$.
If $x$ has guardian $a_k$ at level $\ell{+}1$ and guardian of $a_j$ at level
$\ell$,
then $a_j$ is a type-1 friend of $a_i$, and $d(a_k, a_j) \leq r_\ell +
r_{\ell+1}= \frac12 T_2(\ell)$.  See Figure~\ref{fig:elder}.
\label{thm:elders}
\end{thm}
Based on this scenario, we say $a_j \in A_{\ell}$ is an \emph{elder} of $a_k
\in A_{\ell+1}$ if $d(a_k, a_j) \leq \frac12T_2(\ell)$.
Moreover, if $a_i = a_k$ is its own predecessor, then $a_j = a_i = a_k$ is its
only elder, so the notion of elder is nontrivial only if $a_k$ belongs to
cohort $\ell$.

\begin{thm}
For the purpose of computing elders, there are three particularly good choices
of $\theta$ which require no additional computation.
If $\theta = \frac12(\sqrt{5}-1)$, then $\frac12
T_2(\ell) = T_1(\ell+1)$.
If $\theta = \frac12$, then $\frac12 T_2(\ell) = T_2(\ell+1)$.
If $\theta = \sqrt{2}-1$, then $\frac12 T_2(\ell) = T_3(\ell+1)$.
In these cases, the elders of $a_k$ are the friends (of the respective type)
$a_j$ that have cohort $\ell$ or less.  See Figure~\ref{fig:radii}.
\label{thm:ratios}
\end{thm}

These facts allow the following algorithm.

\subsection{Algorithm}\label{sec:Falgorithm}
At level $0$, the cover tree has a single adult, $a_0$.  All points are its
children.  Its only friend is itself.  Every $x \in \underline{\mathcal{X}}$ has some weight and some
label, which we assume as given input.  For each label $\lambda$, compute the total weight and weighted mean of all children of
$a_0$ with that label.

Suppose at level $\ell$, the friends, children, per-label weights, and per-label means are
known for all adults $a_i \in A_\ell$.  We compute these for level $\ell+1$ as
follows:

\begin{itemize}
\item Advance. Shrink the radius by a factor of $\theta$ to $r_{\ell+1} =
r_\ell \theta$.   Every adult $a_i \in A_\ell$ remains an adult $a_i \in
A_{\ell{+}1}$.  Mark $a_i$ as its own elder.

\item Orphan. For each $a_i \in A_\ell$, find its orphaned children: those $x$
such that $d(a_i, x) > r_{\ell+1}$.  Sort the orphans so that the first orphan
is the one nearest the per-label mean for the label of greatest weight, the
second orphan is the one nearest the per-label mean of the label of
second-greatest weight, and so on, via a two-dimensional argsort without
repetition.  The reason for this sorting will become apparent in
Section~\ref{sec:cder}.

\item Adopt or emancipate. For each $a_i \in A_\ell$, consider its orphans in
order.  For each $x$ orphaned by $a_i$, check whether $x$ can
be adopted by another adult using the type-1 friends of $a_i$:
\begin{itemize}
\item If so, the adoptive guardian will be a adult $a_h$ such that $d(a_h, x)
\leq r_{\ell+1}$.  The predecessor $a_j$ of $a_h$ has $d(a_j, a_i) \leq T_1(l)$.

\item If not, then $x$ is a new adult at level $\ell+1$.  Its only child is
itself.  Its friends are initialized to be itself and its predecessor $a_i$ (but
these will be updated in the Befriending stage).  Note the set $A_{\ell+1}$
is ordered by cohort, and within a cohort is ordered by predecessor, and within a
predecessor block is ordered by proximity to the predecessor's per-label means.
\end{itemize}

\item Exchange Teens.\footnote{This step is optional for general cover
tree, but it is particularly useful for the computation of elders, which are
needed in Section~\ref{sec:cder}.}
If $x$ has guardian $a_i$, then check whether it is closer to some other
guardian using type-2 friends.  That is, find $a_h = \argmin_{A_{\ell+1}} d(x, \cdot)$.
The predecessor $a_j$ of $a_h$ has $d(a_j, a_i) \leq T_2(\ell)$.  Note that
$a_h \neq a_i$ implies $d(a_h, x) > \frac12 r_{\ell+1}$, meaning that only teens (not
youngins) can be exchanged.

\item Befriend: Now, all $(\ell+1)$-adults and their children are known.  We need to
find new friends using the recursion relation and the predecessors.
\begin{itemize}
    \item A pair of $\ell+1$ adults, $a_k$ and $a_h$, are type-3
friends if their predecessors $a_i$ and $a_j$ are type-3 friends at level $\ell$
and $d(a_k, a_h) \leq T_3(\ell+1)$.

    \item A pair of $\ell+1$ adults, $a_k$ and $a_h$, are type-2 friends if they are
type-3 friends and $d(a_k, a_h) \leq T_2(\ell+1)$.

    \item A pair of $\ell+1$ adults, $a_k$ and $a_h$, are type-1 friends if they are
type-2 friends and $d(a_k, a_h) \leq T_1(\ell+1)$.

    \item For each newly emancipated adult $a_k$ of cohort $\ell+1$, find its
    elders by searching its friends for elements of cohort $\ell$ or less.
\end{itemize}
\item Weigh: For each adult $a_k$ at level $\ell+1$, compute the per-label
weights and per-label weighted means for its children.
Sort the labels by their weights from greatest to least.
\end{itemize}
At level $\ell{+}1$, the friends, children, per-label weights, and per-label
means are all known.

\subsection{Algorithmic Efficiency}\label{subsec:eff}
Suppose $n = |\underline{\mathcal{X}}|$, and the cover-tree filtration stops at
level $\ell_{\text{max}}$.  We are interested in the storage and computational
complexity of this algorithm in terms of $n$, for a \emph{fixed} ambient
dimension $D$.

The desired output is the successor/predecessor tree and the children/guardian
relationships at each level.  Thus, storage of friend lists is the only issue
that might lead us to another algorithm.
The main question is: given an adult $a_i$ at level $\ell$, how many other
adults are within $T_3(\ell) = \frac{2}{(1-\theta)}r_\ell$ of $a_i$?

Since no two adults at level $\ell$ are within $r_\ell$ of one another, this
is related to the idea of ``doubling dimension.''   Define $\mathbb{D}(D, \alpha)$
to be the maximum number of points in $\mathbb{R}^D$ that can be placed in the
ball of radius $\alpha$, including the center, so that no two points have
distance $\leq 1$.  This number grows with the ambient dimension $D$.  In
Euclidean space, it is independent of scale; that is, the same number works for
a ball of radius $r\alpha$ using points of distance at least $r$.  The number
can be bounded by the number of simplices with edge length at least $1$ that
can be put into a ball of radius $\alpha$.  Since these simplices have a
minimum volume $\frac{\sqrt{D+1}}{D!2^{D/2}}$, we get a bound like:
\[
\mathbb{D} \leq
\Vol(B_{\alpha})\frac{D! 2^{D/2}}{\sqrt{D+1}}
\sim
\frac{D! (\alpha)^D (2 \pi)^{D/2}}{(\frac{D}{2})!\sqrt{D+1}}
\sim O(D! 2^D \alpha^D)
.\]

Setting $\alpha = \frac{2}{1-\theta}$, we get an upper bound on the expected
size of the friends list, which is a constant independent of the level $\ell$.
Thus, the total storage size is $O(n)$; in particular, it is at most $3
\mathbb{D}n$, where the $3$ accounts for the lists of type-1, type-2, and
type-3 friends.

Setting $\alpha = \frac{1}{\theta}$, we get an upper bound on the expected
number of successors of an adult from level $\ell$ to $\ell+1$, which is a
constant independent of the level $\ell$.  The size of $A_0$ is $|A_0|=1$. The
maximum expected size of $A_1$ is $\mathbb{D}$.  The maximum expected size of
$A_2$ is $\mathbb{D}^2$.  Generally, the maximum expected size of $A_\ell$ is
$\mathbb{D}^\ell$.  Moreover, the larger the size of each cohort $A_\ell \-
A_{\ell-1}$, the shallower the cover tree.  That is, $\ell_{\text{max}}$ is
expected to be of the order $\log_{\mathbb{D}}(n)$.

For each adult $a_i \in A_\ell$, we must:
\begin{enumerate}
\item Sum and sort the weights of its children, by label.  The expected number of
children is $n/\mathbb{D}^D$.  Sorting the labels is $O(L \log(L))$ for $L$ labels.

\item Compute the distance (an $O(D)$ operation) to each of its roughly
$\theta^\ell n$ children. These distances yield the orphans and teens.  The
expected number of orphans is $(1-\theta^D)\frac{n}{\mathbb{D}^\ell}$, and the expected
number of teens is $\left( 1 - (\frac{1}{2})^D\right)\frac{n}{\mathbb{D}^\ell}$.

\item For each orphan, determine whether it should be
adopted or liberated.  Compute the distance (an $O(D)$ operation) to the $T_1$
friends or successors
of the $T_1$ friends of $a_i$.  This set has expected size $\mathbb{D}^2$,
which is constant in $n$.

\item For each teen, determine whether it should be
exchanged.  Compute the distance (an $O(D)$ operation) to the $T_2$
friends or successors of the $T_2$ friends of $a_i$.  This set has expected size $\mathbb{D}^2$,
which is constant in $n$.

\item For each successor---a list of size $\mathbb{D}$---update its friends by
computing the distance (an $O(D)$ operation) to the successors of the $T_3$
friends of $a_i$.  This is expected to be $\mathbb{D}^2$ distance computations.
\end{enumerate}

For each adult at level $\ell$, this is roughly $\frac{n}{\mathbb{D}^\ell}$
operations, albeit with an additional penalty from $D$.  Level $\ell$ has at most $\mathbb{D}^\ell$ adults, in
which case the total number of computations is roughly
\[
\sum_{\ell = 0}^{\ell_{\text{max}}} \mathbb{D}^\ell \frac{n}{\mathbb{D}^\ell}
O(\mathbb{D}^2)
\sim O(n \log_\mathbb{D}n)O(\mathbb{D}^2).
\]

It is interesting that the rapid growth from the ambient dimension $D$ is
counteracted by the logarithm term.   In our experience with applications, this
is effectively linear in $n$, as the sum terminates earlier than this estimate.
This is because the estimate of $\ell_{\text{max}}$ by $\log_\mathbb{D}(n)$ is
typically irrelevant; the maximum level $\ell_{\text{max}}$ of the cover tree
is actually controlled by the ratio of the maximum to the minimum distances in
the pointcloud.  In applications, these are typically within several orders of
magnitude, so with $\theta = \frac12$, we expect $\ell_\text{max} \sim 14$.
Moreover, in our supervised-learning application in Section~\ref{sec:cder}, we
abort the cover tree much earlier, once the low-entropy regions have been
found.

\section{Entropy Reduction}\label{sec:entropy}
For each label $\lambda$, let $\underline{\mathcal{X}}|_\lambda$ denote the
(weighted) points in $\underline{\mathcal{X}}$ that have label $\lambda$.  For
any convex, compact $\Omega$, let $\Omega_\lambda = \Omega \cap
\underline{\mathcal{X}}|_\lambda$.  The total weight in $\Omega$ of each label is
$w_\lambda = \sum \{ w(x) ~:~ x \in \Omega_\lambda \}$, which by
Section~\ref{sec:weight} corresponds to the relative likelihood of a point with
label $\lambda$ being sampled from $\Omega$.
Let $W = \sum_\lambda w_\lambda$.

To quantify the relative prominence of the various labels, we use
the information-theoretic notion of \emph{entropy}.
Entropy is characterized by several nice properties, detailed in \cite{Baez}.
The formula is
\begin{equation}
\begin{split}
S(\Omega) &=
- \sum_{\lambda} \frac{w_\lambda}{W}\log_L\left(
\frac{w_\lambda}{W}\right)\\
&=
-\frac{1}{W}\sum_{\lambda} w_\lambda\log_L\left(
w_\lambda\right) + \log_L(W).
\end{split}
\end{equation}

If all weights are roughly equal, then entropy is 1.  If one label has high
weight and the others have weight 0, then the entropy is 0.  Generally, entropy
decreases as a smaller subset of labels becomes more likely.  Thus, low entropy
is a good indicator that the region $\Omega$ is particularly prominent for a
small subset of labels.
The question remains: how can we locate low-entropy regions?

Our search for low-entropy regions is motivated by the following elementary
theorems; however, we do not insist that these strict regularity hypotheses
hold for any given application.

\begin{thm}
Fix a function $f:\mathbb{R}^D \to \mathbb{R}$ that is continuous on
a compact, convex set $\Omega$.
Fix a radius $r>0$.  Let $\mu_r(x)$ denote the centroid of $f$ on the compact,
convex set $\Omega \cap B_r(x)$ (the closed ball).
There exists $x \in \Omega$ such that $\mu_r(x) = x$.
\end{thm}
Proof: Note that $\mu_r(x)$ varies continuously with $x$, and apply Brouwer's fixed-point
theorem (\cite{Munkres1993}, II.22).

\begin{thm}
Suppose that $f:\mathbb{R}^{D} \to \mathbb{R}$ is an analytic function with local
max at $x$.   For any $\epsilon >0$, there exists a radius $r$ such that the
centroid $\mu_r(x)$ of $f$ on $B_r(x)$ has $d(\mu_r(x) ,x ) < \epsilon$.
\end{thm}
Proof: Expand $f$ as a power series around $x$, and approximate $f$ by
truncating at the second-order term.  The level sets are concentric ellipsoids about $x$, as given
by the positive-definite symmetric matrix $f''$.

\begin{thm}
Suppose $f$ and $\tilde{f}$ are positive, continuous functions defined on a
compact, convex region $\Omega \subset \mathbb{R}^D$.
Suppose that $f$ achieves a local max $f(x) = M$ on $\Omega$ and that
$\tilde{f}(x) \leq \varepsilon$ on $\Omega$, with $\varepsilon < M$.  Then,
there is a radius $r$ such that the average value of $f$ on $\Omega \cap
B_s(x)$ is greater than the average value of $\tilde{f}$ on $\Omega \cap
B_s(x)$ for all $s \leq r$.
\end{thm}
Proof:   The mean value theorem for integrals.

Putting these together, we try to find a region $\Omega$ and a ball of radius
$r_\ell$ in that region whose center is near the mean of a particular label.
We ask whether the entropy of that ball is non-decreasing with radius;  that
is, does the entropy become lower (or remain the same) when the ball shrinks?
If true, this is consistent with the hypothesis that a local maximum of a
particular label occurs near the center of that ball, while other labels remain
roughly constant.  If false, then we subdivide the region and search around the
boundary of the original ball for other smaller regions of low entropy.

Specifically, the compact, convex regions corresponds to the children of an
adult $a_i \in A_\ell$ in the cover tree, and the sub-balls correspond to the
children of successors of $a_i$ at level $\ell+1$ in the cover tree.   This is
the heart of the CDER algorithm in Section~\ref{sec:cder}.

\section{Cover-tree Differencing via Entropy Reduction}\label{sec:cder}
The input data is a ratio $\theta$ and a training set of pointclouds
$X_1, X_2, \ldots, X_N$ with labels $\lambda_1, \lambda_2, \ldots, \lambda_N$,
respectively.    Set weights as in Section~\ref{sec:weight}.

\subsection{Region Selection}\label{sec:select}
Here we describe the heart of the CDER algorithm, wherein we search the adults
of the cover tree as a data-driven method to locate regions on which we want to
build distributional coordinates.  The construction and weighting of those
distributional coordinates occurs in Section~\ref{sec:build}.

For each level of the cover tree, define a subset $\hat{A}_\ell \subset A_\ell$
of adults that are still potential candidates for distributional coordinates.
Set $\hat{A}_0 = A_0 = \{ a_0\}$.  Once $\hat{A}_\ell = \{ \}$, we
\texttt{break}.

The cover tree of $\underline{\mathcal{X}}$ is constructed level-by-level as in
Section~\ref{sec:Falgorithm}.  At each level $\ell$ of the cover tree (until \texttt{break}), we
perform the following computation:

For each adult $a_i \in A_\ell$, let $\beta_\ell(i)$ denote the set of children
of $a_i$ at level $\ell$. Let $\alpha_\ell(i) = \beta_{\ell+1}(i)$ denote the
set of children of $a_i$ at level $\ell+1$.  Let $\gamma_\ell(i)$ denote the
union of the children (at level $\ell-1$) of the elders of $a_i$.
(If $a_i$ belongs to cohort $\ell-1$ or less, then $\gamma_\ell(i) =
\beta_{\ell-1}(i)$.)
Note that $\alpha_\ell(i) \subset \beta_\ell(i) \subset \gamma_\ell(i)$; see
Figure~\ref{fig:elders}.  Then, we decide what to do with $a_i$:

\begin{figure}
\begin{center}
\begin{tikzpicture}[scale=0.3]
\fill[opacity=0.4,color=yellow] (-13,0) circle (4.5cm);
\fill[opacity=0.4,color=orange] (-13,0) circle (3cm);
\fill[opacity=0.4,color=red] (-13,0) circle (2cm);
\fill (-13,0) circle (0.1cm);
\node [right] at (-13,0)  {$a_i$};
\node [below] at (-13,2)  {$\alpha$};
\node [below] at (-13,3.25)  {$\beta$};
\node [below] at (-13,4.5)  {$\gamma$};
\fill[opacity=0.4,color=yellow] (3.1,0) circle (4.5cm);
\fill[opacity=0.4,color=yellow] ({4*cos(120)},{4*sin(120)}) circle (4.5cm);
\fill[opacity=0.4,color=yellow] ({3.5*cos(240)},{3.5*sin(240)}) circle (4.5cm);
\fill[opacity=0.4,color=orange] (0,0) circle (3cm);
\fill[opacity=0.4,color=red] (0,0) circle (2cm);
\fill (0,0) circle (0.1cm);
\fill (3.1,0) circle (0.1cm);
\fill ({4*cos(120)},{4*sin(120)}) circle (0.1cm);
\fill ({3.5*cos(240)},{3.5*sin(240)}) circle (0.1cm);
\node [above] at ({4*cos(120)}, {4*sin(120)}) {\small{elder}};
\node [right] at (0,0)  {$a_i$};
\node [below] at (0,2)  {$\alpha$};
\node [below] at (0,3.25)  {$\beta$};
\node [below] at (0,4.5)  {$\gamma$};
\end{tikzpicture}
\end{center}
\caption{The nested sets $\alpha_\ell(i) \subset \beta_\ell(i) \subset
\gamma_\ell(i)$ on which we compare entropy. On left, $a_i$ has cohort $<\ell$
and is its own elder.  On right, $a_i$ has cohort $\ell$, and has several
elders as in Theorem~\ref{thm:elders}. If the smaller regions have lower entropy, then we
build a distributional coordinate using $a_i$ at level $\ell$.}\label{fig:elders}
\end{figure}
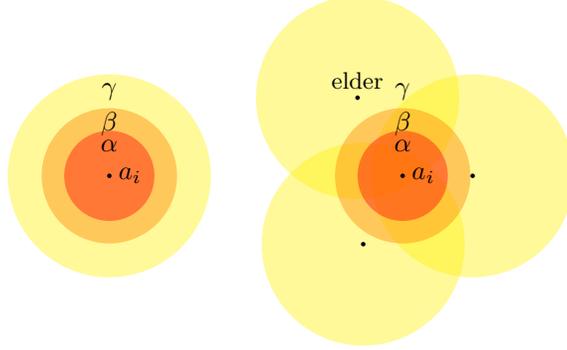

\begin{itemize}
\item if $\hat{A}_\ell = \emptyset$, then \texttt{break}.  Nothing more is needed.
\item for each adult $a_i \in \hat{A}_\ell$:
\begin{itemize}
\item if $a_i$ has only one child (itself) at level $\ell+1$, then \texttt{pass}.  This
    region is useless.
\item elif $a_i$ has only one child (itself) at level $\ell$, then \texttt{pass}.  This
    region is useless.
\item elif $S(\alpha) \leq S(\beta) \leq S(\gamma) < 1$, then
    \texttt{build} a distributional coordinate using the children of $a_i$ of
    each of the dominant labels.\footnote{A label is dominant if its points represent more
    than $\frac1L$ of the total weight of the children.  For a two-label task,
    this is simply the majority label.}
\item  elif $S(\gamma) \leq S(\alpha) \leq S(\beta) < 1$, then \texttt{append} $a_i$ to
    $\hat{A}_{\ell+1}$.  The smaller ball $B(a_i, r_{\ell+1})$ may contain a
    lower-entropy region.
\item elif $S(\beta) \leq S(\alpha) \leq S(\gamma) < 1$, then \texttt{append} the successors of
    $a_i$---other than itself!---to $\hat{A}_{\ell+1}$.  The annulus $B(a_i,
    r_\ell) \setminus B(a_i, r_{\ell+1})$ may contain a lower-entropy region.
\item elif $S(\beta) \leq S(\gamma) \leq S(\alpha) < 1$, then \texttt{append} the successors of
    $a_i$---other than itself!---to $\hat{A}_{\ell+1}$. The annulus $B(a_i,
    r_\ell) \setminus B(a_i, r_{\ell+1})$ may contain a lower-entropy region.
\item elif $S(\gamma) \leq S(\alpha) \leq S(\beta) < 1$, then \texttt{pass}.  The region $B(a_i,
    r_\ell)$ is useless.
\item elif if $S(\gamma) \leq S(\beta) \leq S(\alpha) < 1$, then \texttt{pass}. The region $B(a_i,
    r_\ell)$ is useless.
\item else (meaning $\max\{ S(\alpha), S(\beta), S(\gamma)\} = 1$) \texttt{append} the
    successors of $a_i$ to $\hat{A}_{\ell+1}$.  There is too little
    information to decide whether $B(a_i, r_\ell)$ contains a low-entropy
    region, so re-analyze everything at the next level.
\end{itemize}
\end{itemize}

Recall that our method in Section~\ref{sec:Falgorithm} produces the list
$A_{\ell}$ which is ordered by cohort, and within a cohort is ordered by
predecessor, and within a predecessor block is ordered by proximity to the
predecessor's per-label means.  Therefore, if $a_i < a_j$ in this ordering,
then $a_i$ tends to be nearer the mean of the most prominent label of a larger
region than $a_j$.  Hence, the distributional coordinates are sorted by
``granularity,'' from coarse to fine.

In some scenarios, detailed analysis of clusters is more important than
efficient classification.  Therefore, one can implement a ``non-parsimonious''
version by replacing all \texttt{pass} operations with \texttt{append}.  For
machine learning applications, the ``parsimonious'' version is far faster, as
the cover tree can be stopped much earlier than the limit suggested in
Section~\ref{subsec:eff}.

\subsection{Build Distributional Coordinates}\label{sec:build}
For each $a_i \in A_{\ell}$ selected to \texttt{build}, we construct a
distributional coordinate $g_i$ in the following way.

For each of the dominant labels $\lambda$ among the children of $a_i$, use
PCA/SVD to build a Gaussian model $g_{i,\lambda}$ of mass 1.   Let $\Delta S$
denote the entropy difference on $\beta_\ell$ caused by erasing the
non-dominant labels.
We amplify (or attenuate) each
Gaussian model by the
coefficient
\[
m_{i} = w_\lambda  (1 - \Delta S) \left(r_\ell\right)^D.
\]
The distributional coordinate is $m_i g_{i,\lambda}(x)$.

One can experiment with different weighting systems; we have found these to be
generally effective, and we justify this choice for the following geometric
reasons:
\begin{itemize}
\item The weight represents the relative likelihood that a point of the given
label was selected from the region on which this coordinate was detected.
Perhaps, by chance in our selection process, a single region was detected at a
later level in the form of several smaller regions.  The weight of the smaller
regions will sum to the weight of the larger region.

\item The entropy term is the ``certainty'' with which this region was
selected.  The entropy $\Delta S$ is the information that is \emph{lost} if the
other labels are removed from the region, so $1-\Delta S$ is the information
that remains. In other words, this term penalizes regions according to the
impurity of their labels.

\item The $(r_\ell)^D = (r_0 \theta^\ell)^D$ term accounts for the relative
remoteness of the sample region.  All else being equal, remote regions are
particularly distinctive. Because of the cover-tree construction, remoteness
can be measured by the size of the cover-tree ball.
For example,  suppose that two different
regions are detected by CDER, each with the same entropy, weight, and
configuration of points; the Gaussian coordinates are translates of
one-another.  Suppose that the first region is very distant from the rest of
the pointcloud, so it is detected at an early level $\ell$.  Suppose that the second region
is surrounded closely by other data, so it is not detected by the cover tree
until a later level $\ell' > \ell$. The first region's relative volume is
$\theta^{(\ell-\ell')D}$.
\end{itemize}

\section{Examples}\label{sec:examples}
This section illustrates the promise of CDER on several synthetic examples
of varying levels of complexity that are designed to stress certain features of
the algorithm.

\subsection{Blobs}\label{ex:blobby}
This example shows that CDER ignores high-density regions unless they
have low entropy.

\begin{figure}
\begin{center}
\includegraphics[width=0.4\columnwidth]{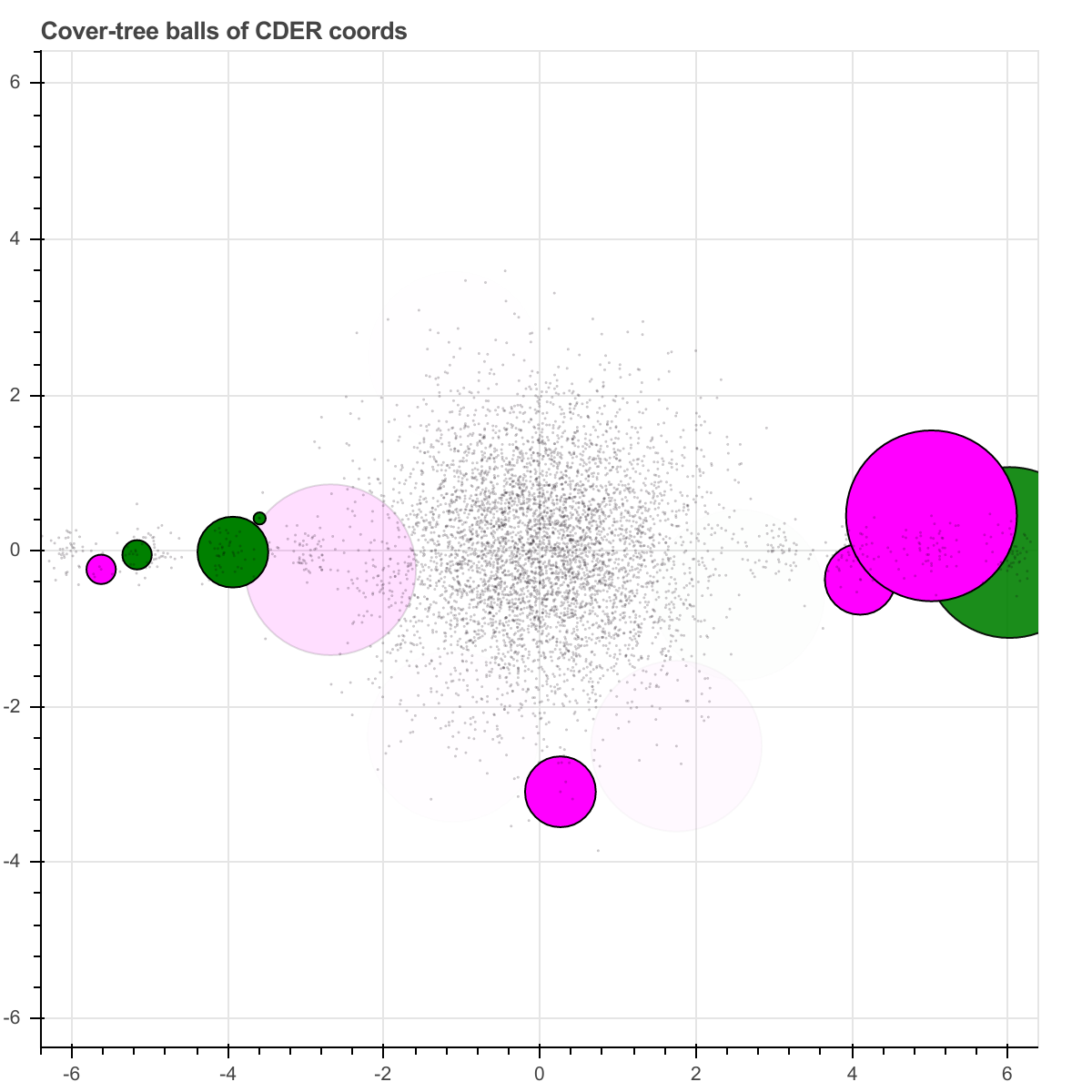}
\includegraphics[width=0.4\columnwidth]{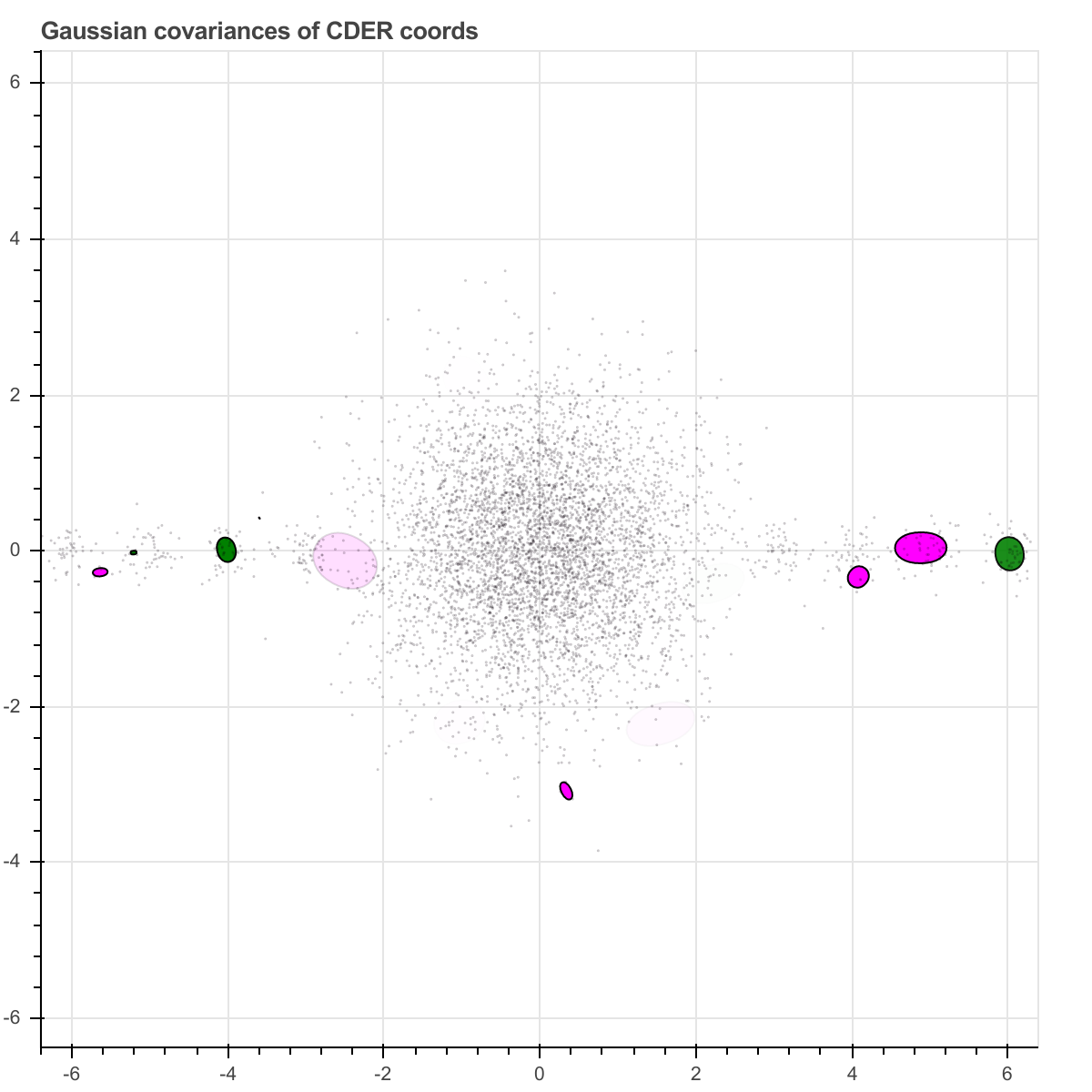}\\
\includegraphics[width=0.8\columnwidth]{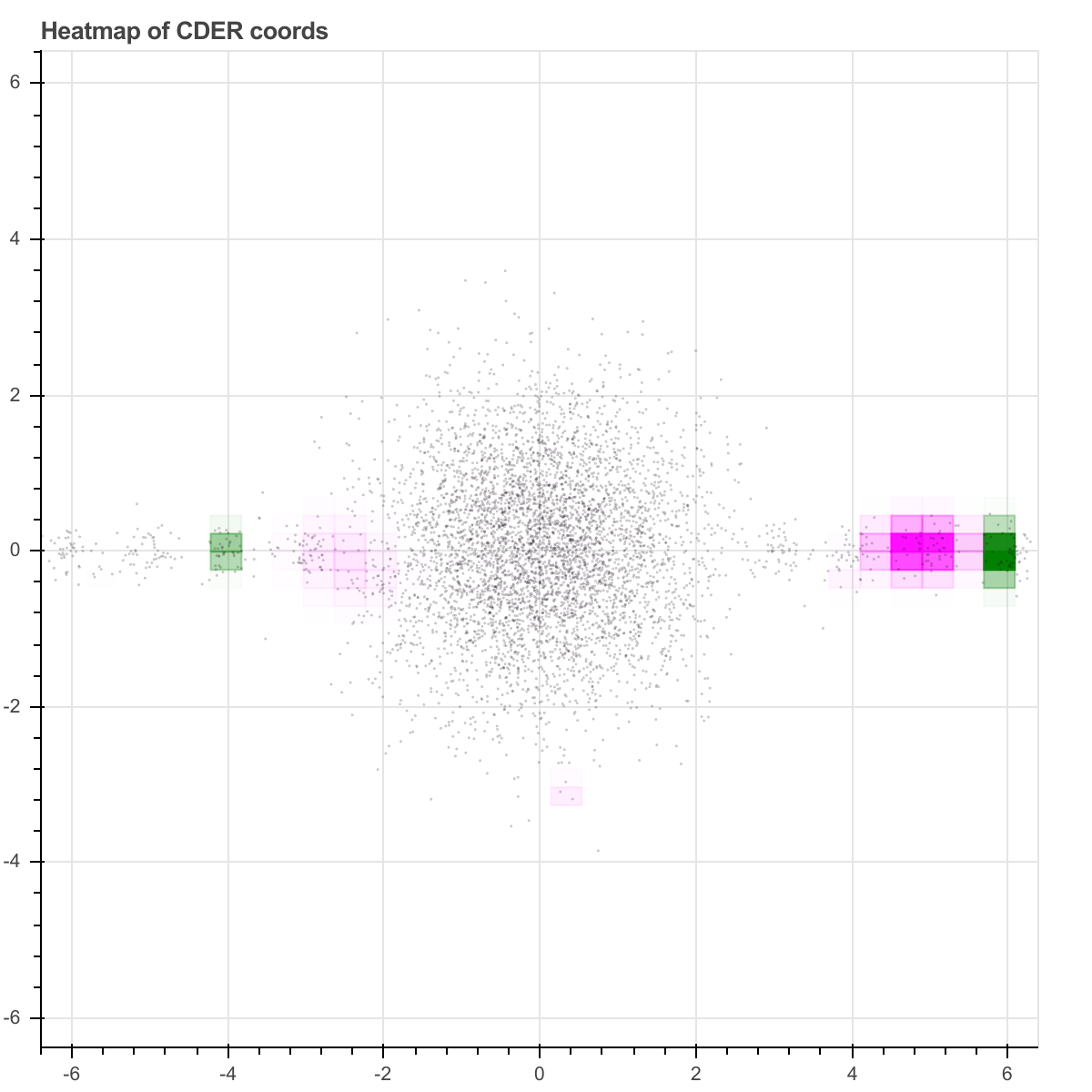}
\end{center}
\caption{The output of CDER for the example of
Figure~\ref{fig:blobby}.  Regions are shaded by certainty (1-entropy).}\label{fig:blobbyout}
\end{figure}

Consider two labels\footnote{Labels are indicated by colors, and we have intentionally chosen colors that seem to show up non-catastrophically in black-and-white printouts. Nonetheless, we recommend printing this paper in color or at least reading this section on the computer screen!}: 0/magenta and 1/green.
The cloud collection has 25 pointclouds of each label.
Each magenta pointcloud consists of:
\begin{itemize}
\item 100 points sampled from a standard normal
distribution
\item 2 points sampled from a normal distribution
with $\sigma = 0.2$ and mean $(4,0)$
\item 2 points sampled from a normal distribution
with $\sigma = 0.2$ and mean $(5,0)$
\item 2 points sampled from a normal distribution
with $\sigma = 0.2$ and mean $(-3,0)$
\item 2 points sampled from a normal distribution
with $\sigma = 0.2$ and mean $(-6,0)$
\end{itemize}
Each green pointcloud consists of
\begin{itemize}
\item 100 points sampled from a standard normal
distribution
\item 2 points sampled from a normal distribution
with $\sigma = 0.2$ and mean $(-4,0)$
\item 2 points sampled from a normal distribution
with $\sigma = 0.2$ and mean $(-5,0)$
\item 2 points sampled from a normal distribution
with $\sigma = 0.2$ and mean $(3,0)$
\item 2 points sampled from a normal distribution
with $\sigma = 0.2$ and mean $(6,0)$
\end{itemize}
This produces the cloud collection in Figure~\ref{fig:blobby}.
The output of CDER appears in Figure~\ref{fig:blobbyout}

In a typical run, the algorithm terminates at cover-tree level 8, even though
the complete cover tree of $\underline{\mathcal{X}}$ ends at level 14.  A total
of 26 distributional coordinates are produced, and their weights vary by about
3 orders of magnitude.
Table~\ref{table} shows how conservative the entropy-reduction process is at
selecting regions to generate distributional coordinates.

\begin{table}
\begin{tabular}{lrrc}
$\ell$ & $|A_\ell|$ & $|\hat{A}_\ell|$ & new Gaussians\\\hline
0 & 1 & 1 & +0 \\
1 & 7 & 7 & +0 \\
2 & 29 & 29 & +4\\
3 & 119 & 37 & +7\\
4 & 446 & 36 & +7\\
5 & 1379 & 40 & +6\\
6 & 2827 & 23 & +1\\
7 & 3903 & 3 & +1
\end{tabular}
\caption{CDER is parsimonious about selecting cover-tree regions for
distributional coordinates.}\label{table}
\end{table}

To judge accuracy on testing data in a supervised-learning scenario, we need a
method to label a test pointcloud using the distributional coordinates.  Each
test pointcloud $X_{\mathrm{test}}$ is mapped to a point in $\mathbb{R}^{26}$.
Many sophisticated methods are possible, but for simplicity, we simply ask: For a
given pointcloud $X_{\mathrm{test}}$, which is bigger:  the Euclidean norm
of the magenta Gaussian coordinates evaluated on $X_{\mathrm{test}}$,
\[ \sqrt{\sum_{g~\mathrm{magenta}} \left(\int_{X_{\mathrm{test}}} g\right)^2}\]
or the Euclidean norm of the green Gaussian coordinates evaluated on
$X_{\mathrm{test}}$,
\[ \sqrt{\sum_{g~\mathrm{green}} \left(\int_{X_{\mathrm{test}}} g\right)^2}?\]

With this simple comparison, the algorithm achieves 100\% accuracy in a 5-fold
cross-validation of this cloud collection with a 80/20 training/testing split.
More precisely, for each run of the cross-validation, we take 80 percent of the point clouds,
turn these into a cloud collection, build distributional coordinates using this cloud collection, and then
test the results on the remaining 20 percent of the point clouds. This entire procedure is then repeated five times.

Moreover, the relative masses of the 28 distributional coordinates vary over
four orders of magnitude, so for this sort of comparison, one could dispose of
many of them in the interest of speed while preserving accuracy.
Note that these distributions are contrived to have the same mass (0-moment)
and mean (1-moment) and variance.  Elementary statistical tests would not
distinguish them; 2-moments or skewness tests would be necessary.

\subsection{Blocks}\label{ex:blocky}
This example shows that CDER is fairly robust against background noise that
prevents strict separation.  It also demonstrates that smoothness of the
underlying distributions is not necessary for good results.

\begin{figure}
\begin{center}
\includegraphics[width=0.45\columnwidth]{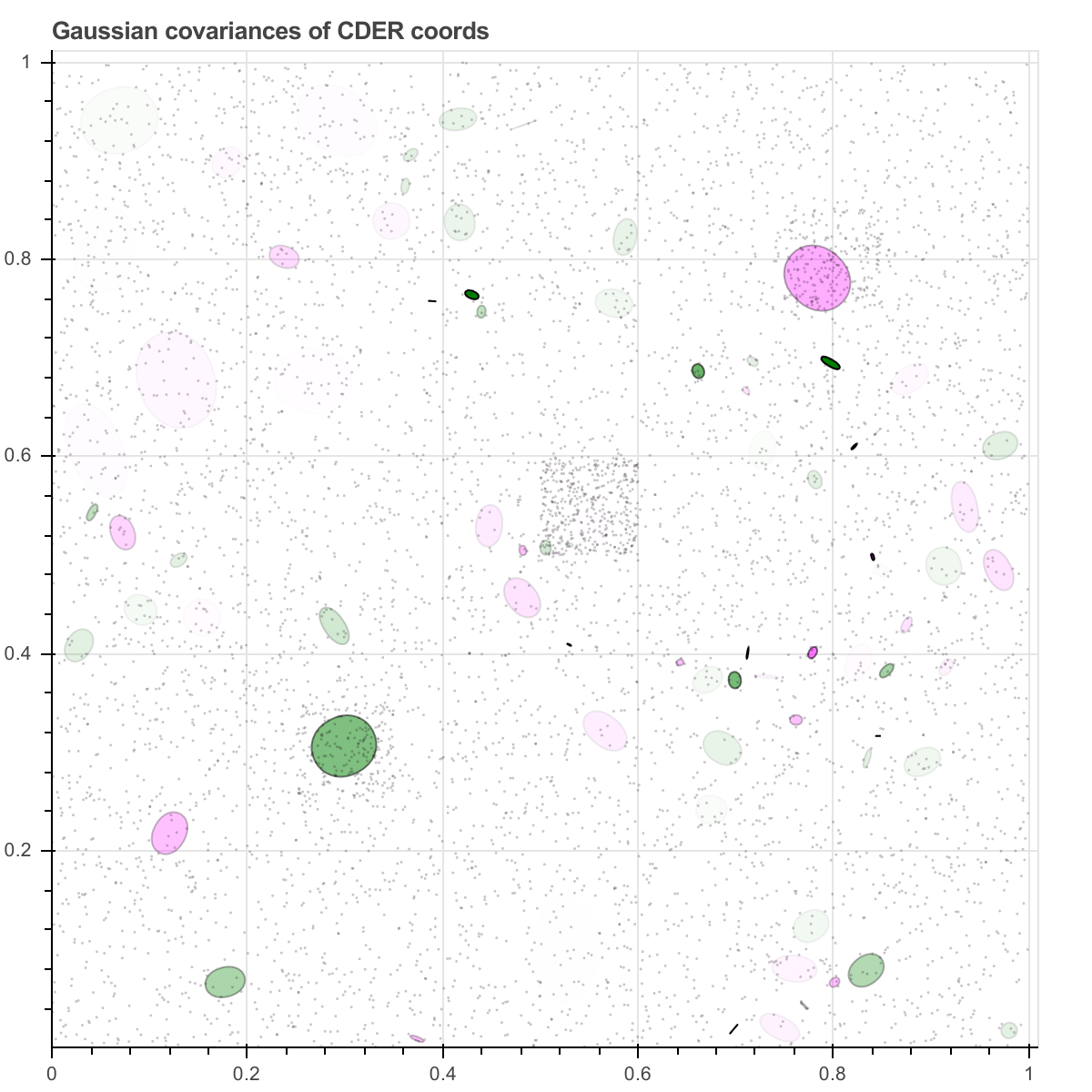}
\includegraphics[width=0.45\columnwidth]{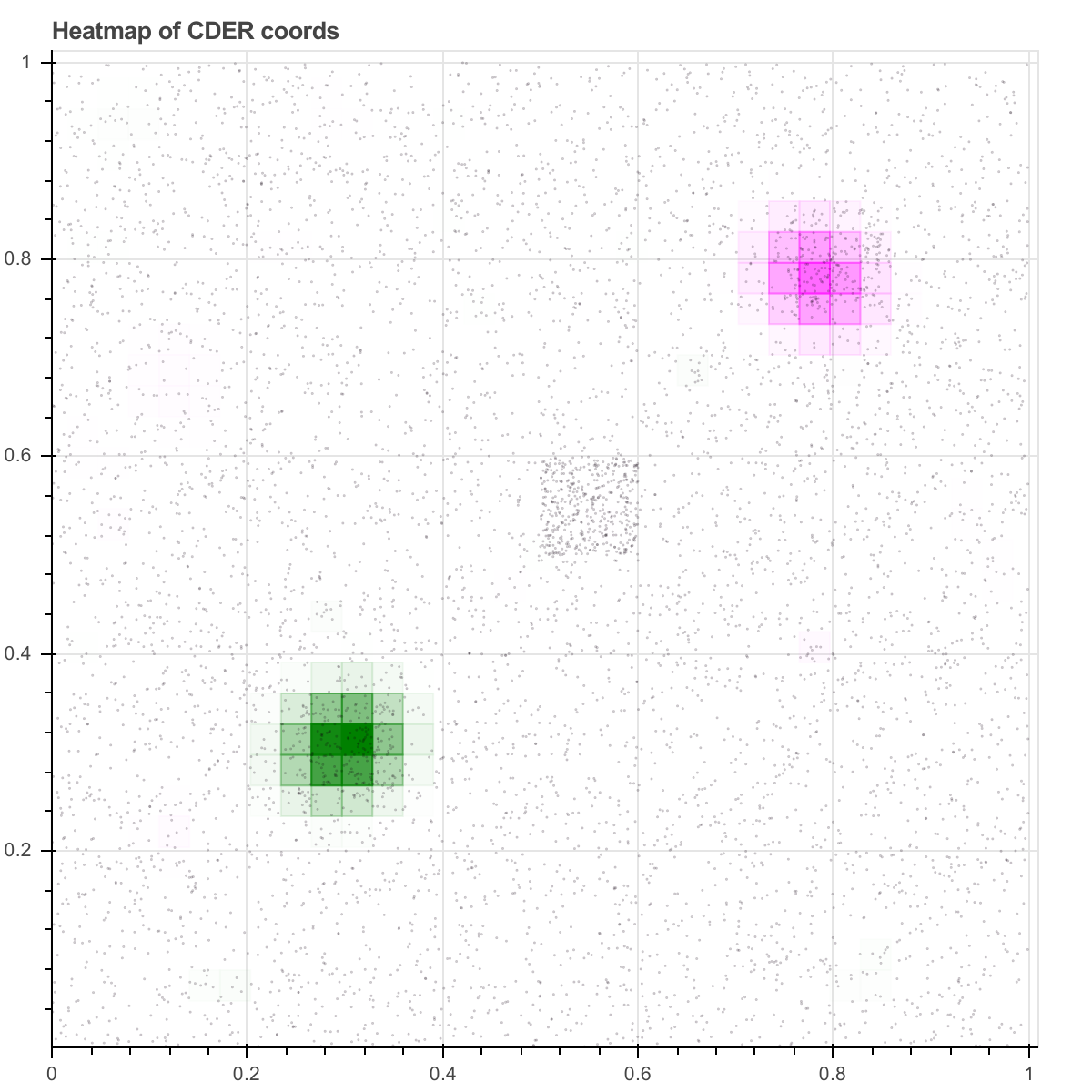}
\end{center}
\caption{The distributional coordinates found by CDER from
Figure~\ref{fig:blocky}.}\label{fig:blockyout}.
\end{figure}

Consider two labels: 0/magenta and 1/green.
The cloud collection consists of 100 magenta pointclouds and 100 green
pointclouds.
Each magenta pointcloud in generated by sampling 30 points uniformly in the
unit square, as well as 2 extra points sampled uniformly in a $0.1\times 0.1$ square at
$(\frac14, \frac14)$, as well as
2 extra points sampled uniformly in a $0.1\times 0.1$ square at
$(\frac12, \frac12)$.
Each green pointcloud in generated by sampling 30 points uniformly in the
unit square, as well as 2 extra points sampled uniformly in a $0.1\times 0.1$ square at
$(\frac34, \frac34)$, as well as
2 extra points sampled uniformly in a $0.1\times 0.1$ square at
$(\frac12, \frac12)$.
See Figures~\ref{fig:blocky} and \ref{fig:blockyout}.
Using the same simple comparison as in Section~\ref{ex:blobby}, the algorithm
achieves 88\% accuracy despite the high background noise.

\subsection{Deep Field}
In this example, we demonstrate that Gaussian mixtures are (nearly)
``fixed points'' of CDER.  It also demonstrates that the algorithm can handle
unequal sample sizes via the weighting system in Section~\ref{sec:weight}.

\begin{figure}
\begin{center}
\includegraphics[width=0.45\textwidth]{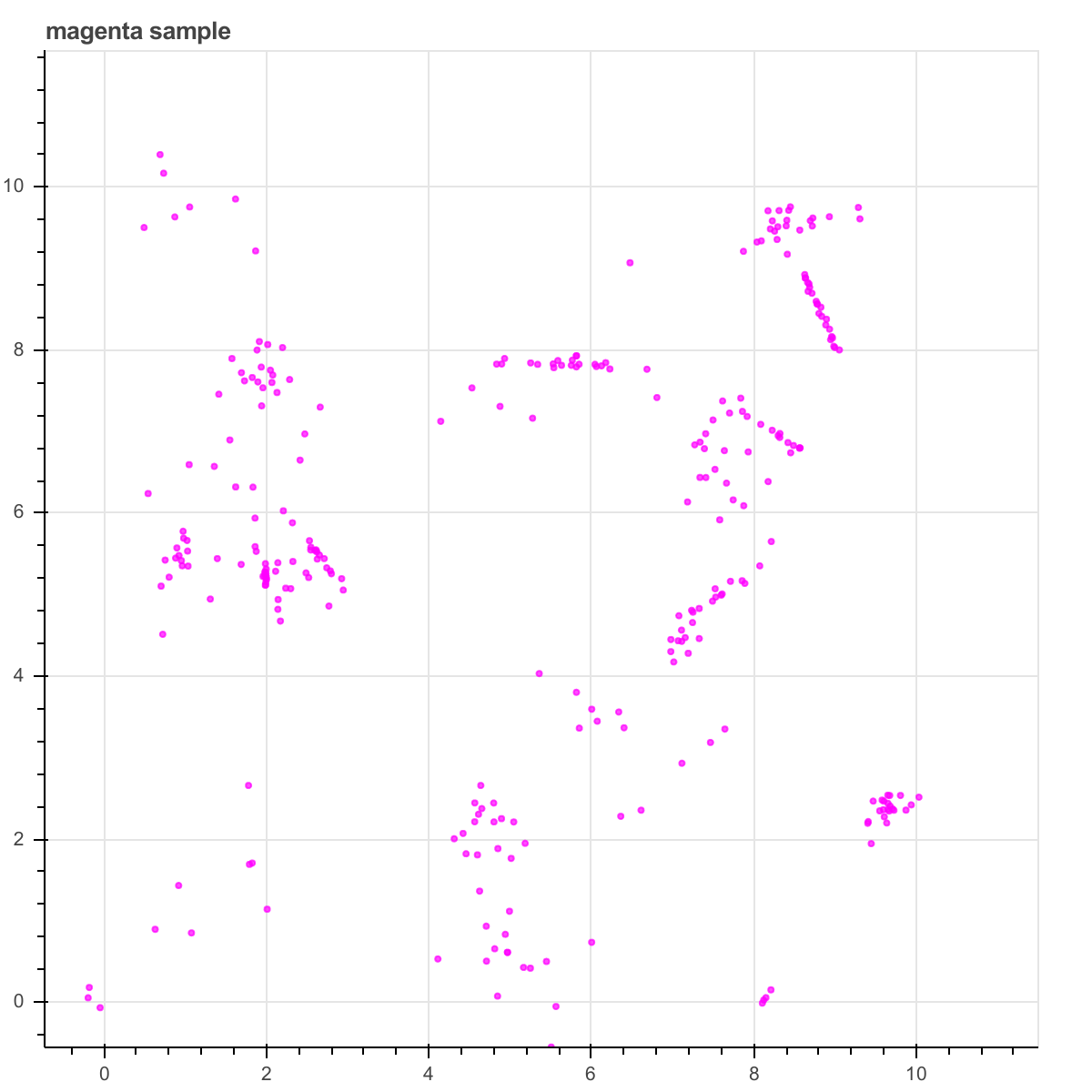}
\includegraphics[width=0.45\textwidth]{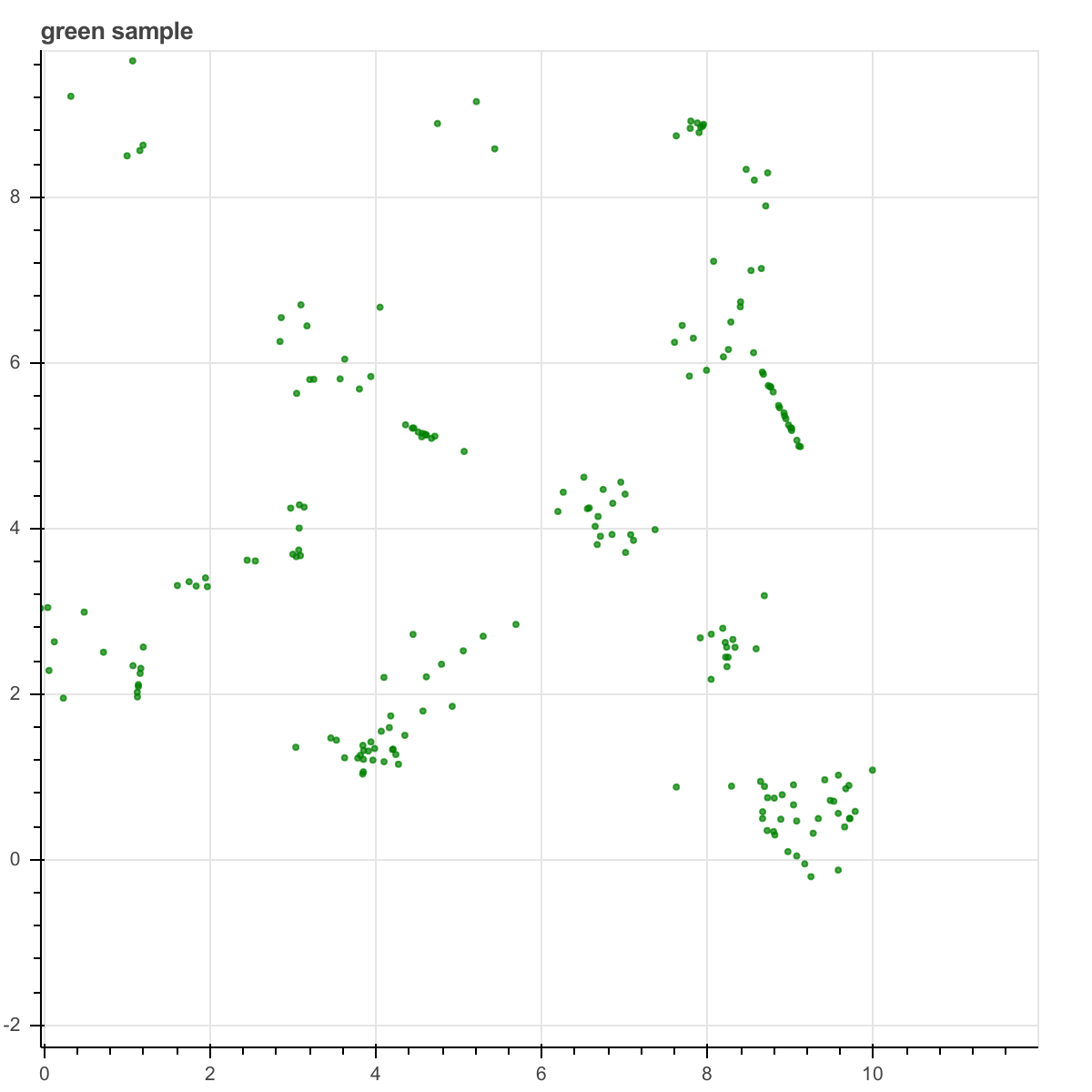}\\
\includegraphics[width=0.70\textwidth]{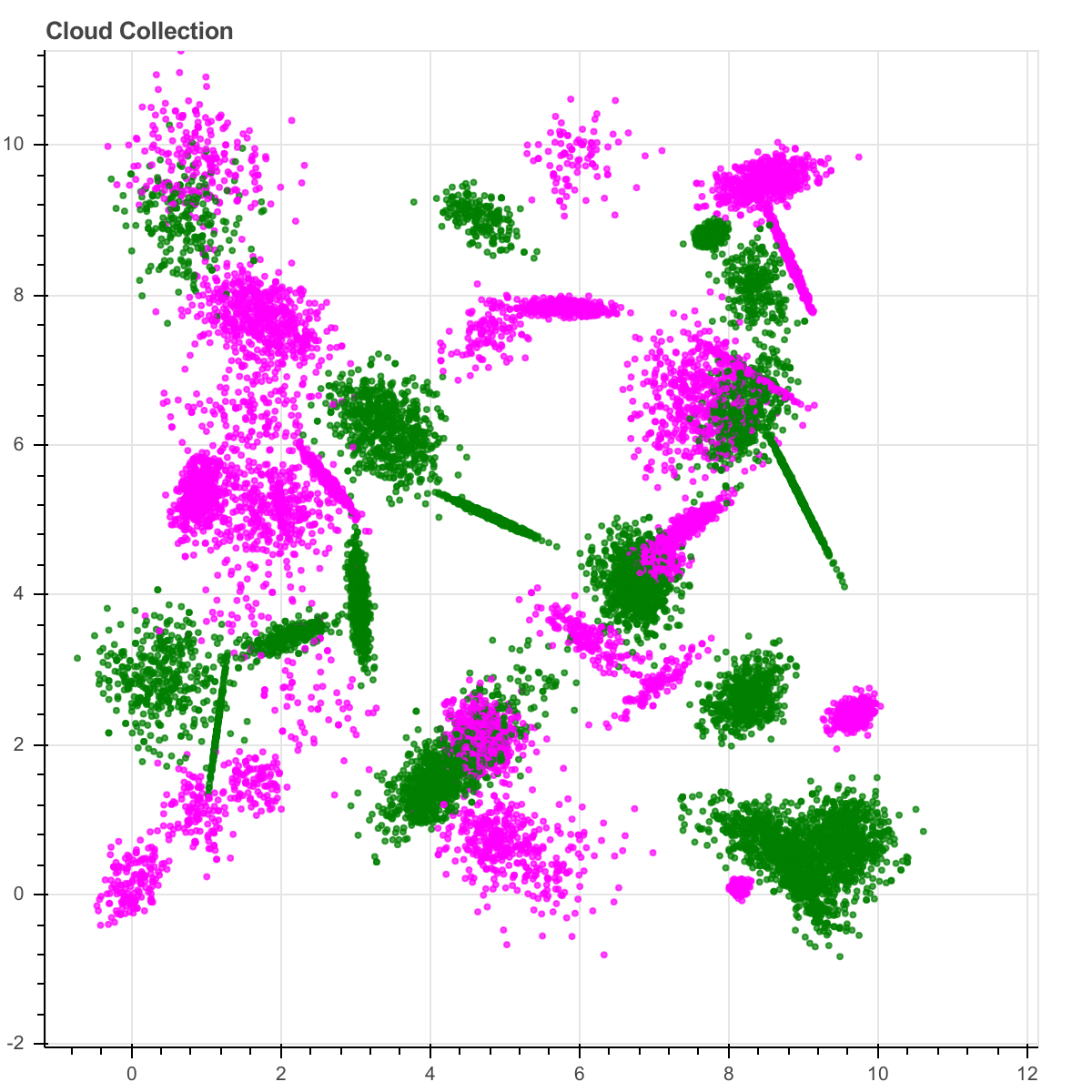}
\end{center}
\caption{A single magenta pointcloud, a single green pointcloud, and the
total cloud collection for a random ``deep field''
example.}\label{fig:deepfield}
\end{figure}

\begin{figure}
\begin{center}
\includegraphics[width=0.45\textwidth]{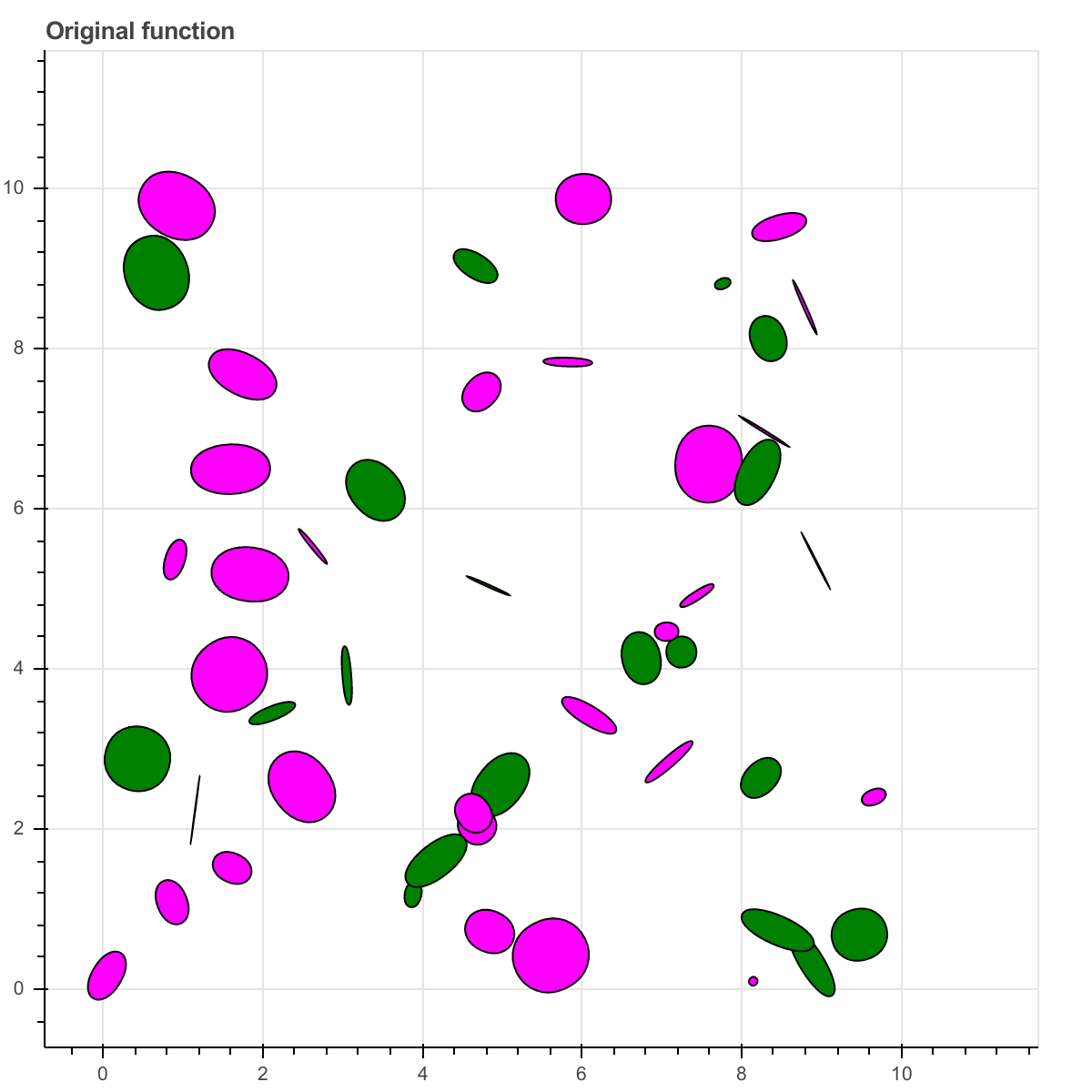}
\includegraphics[width=0.45\textwidth]{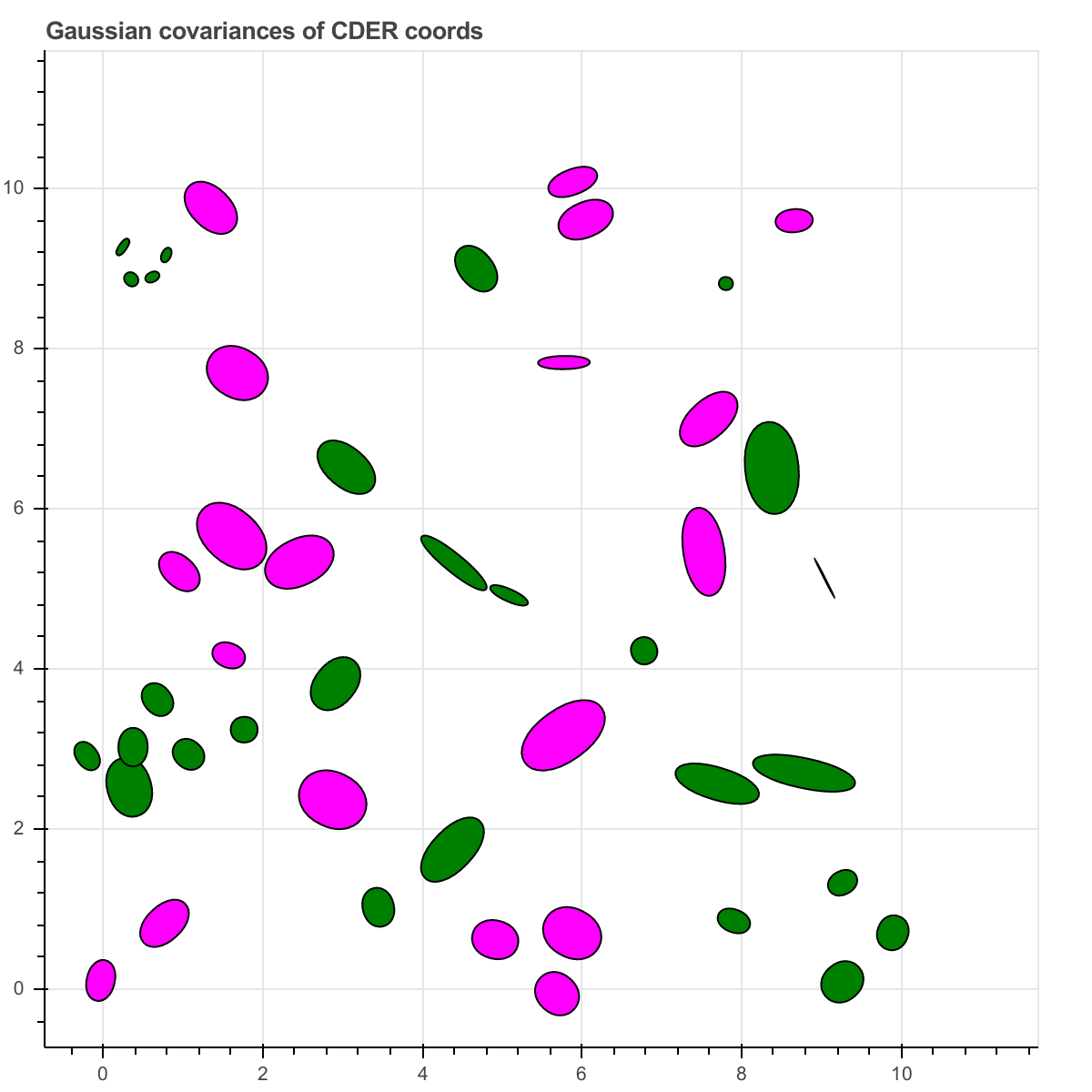}
\end{center}
\caption{Left: The original randomly-generated Gaussian mixture model.  Right:
The Gaussian mixture model returned by CDER.}\label{fig:deepfieldcompare}
\end{figure}

Consider two labels: 0/magenta and 1/green.
The underlying distribution is build from 50 Gaussian distributions.
For each of these, we choose a random label.  We choose a mean point, uniformly
on the $10 \times 10$ square.  We choose a pair of orthonormal covariances uniformly
from $0$ to $0.5$.  We choose a random rotation angle.  We choose an
amplification factor (that is, a sample size) uniformly between 50 and 5000.
For a particular random seed, we arrive at some combined density function $f_0$ and
$f_1$, as in Figure~\ref{fig:deepfieldcompare}

Then, for each label, we produce between 20 and 40 pointclouds,
each with between 50 and 500 points.  For a particular random seed, we arrive
at the cloud collection in Figure~\ref{fig:deepfield}.

\begin{figure}
\begin{center}
\includegraphics[width=0.45\textwidth]{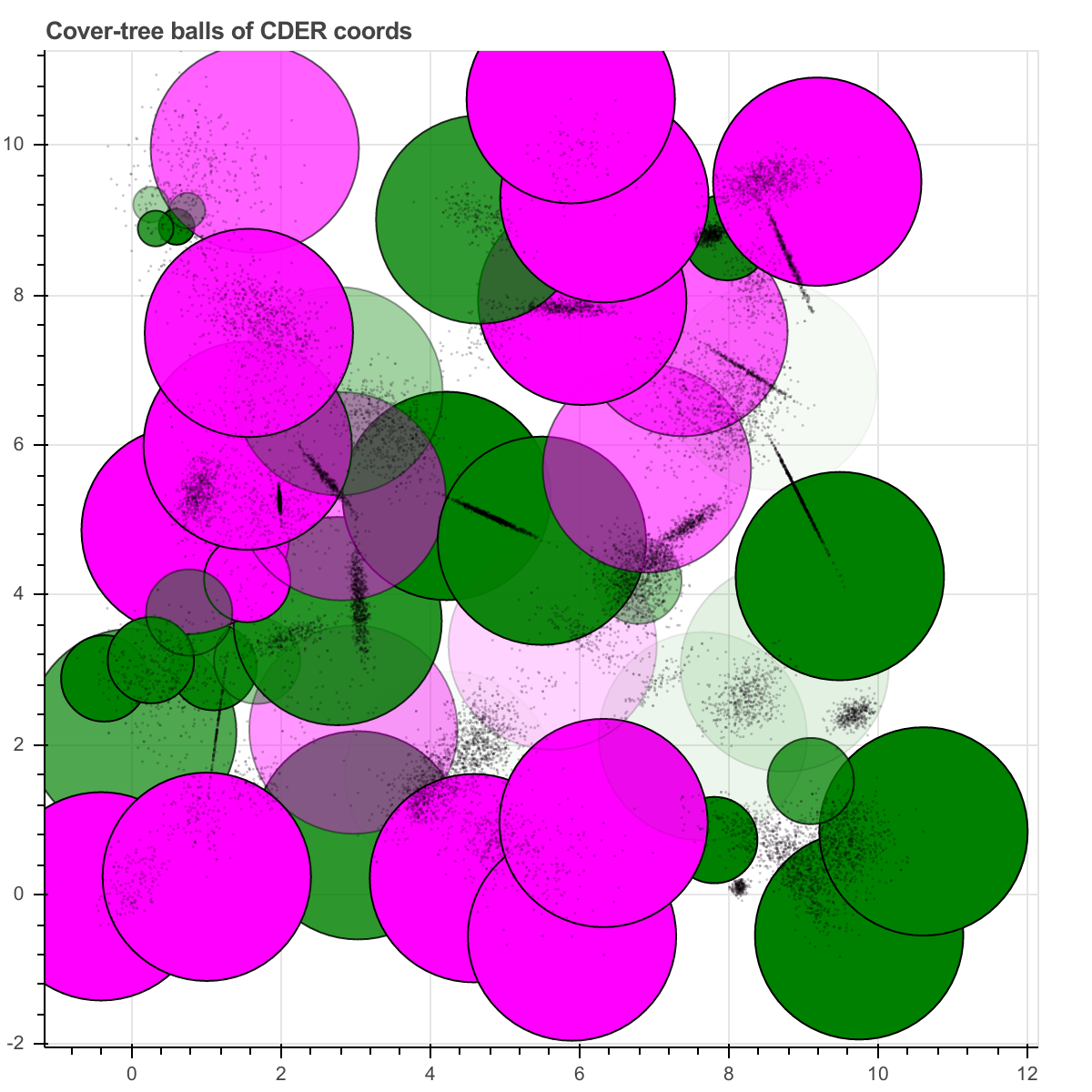}
\includegraphics[width=0.45\textwidth]{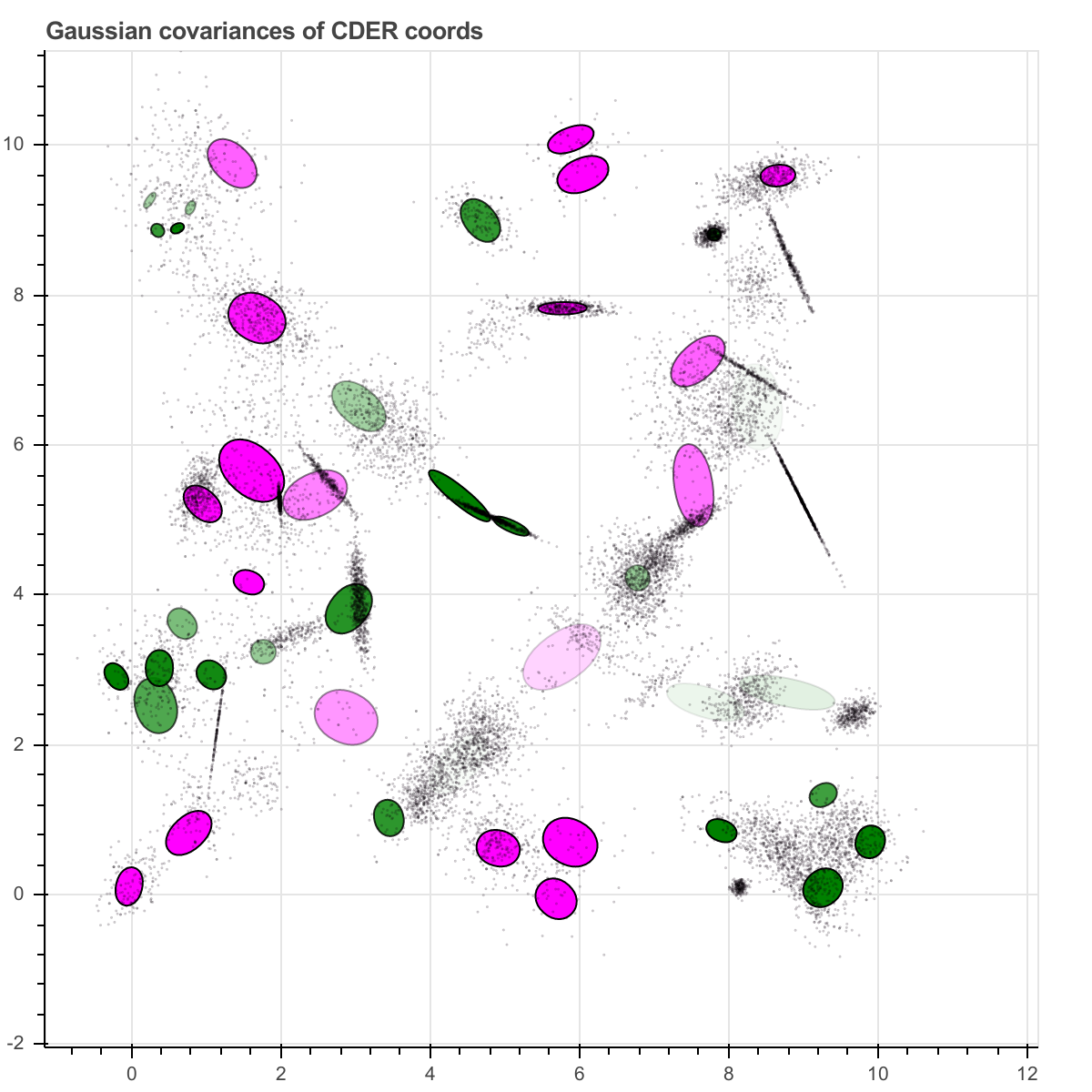}\\
\includegraphics[width=0.90\textwidth]{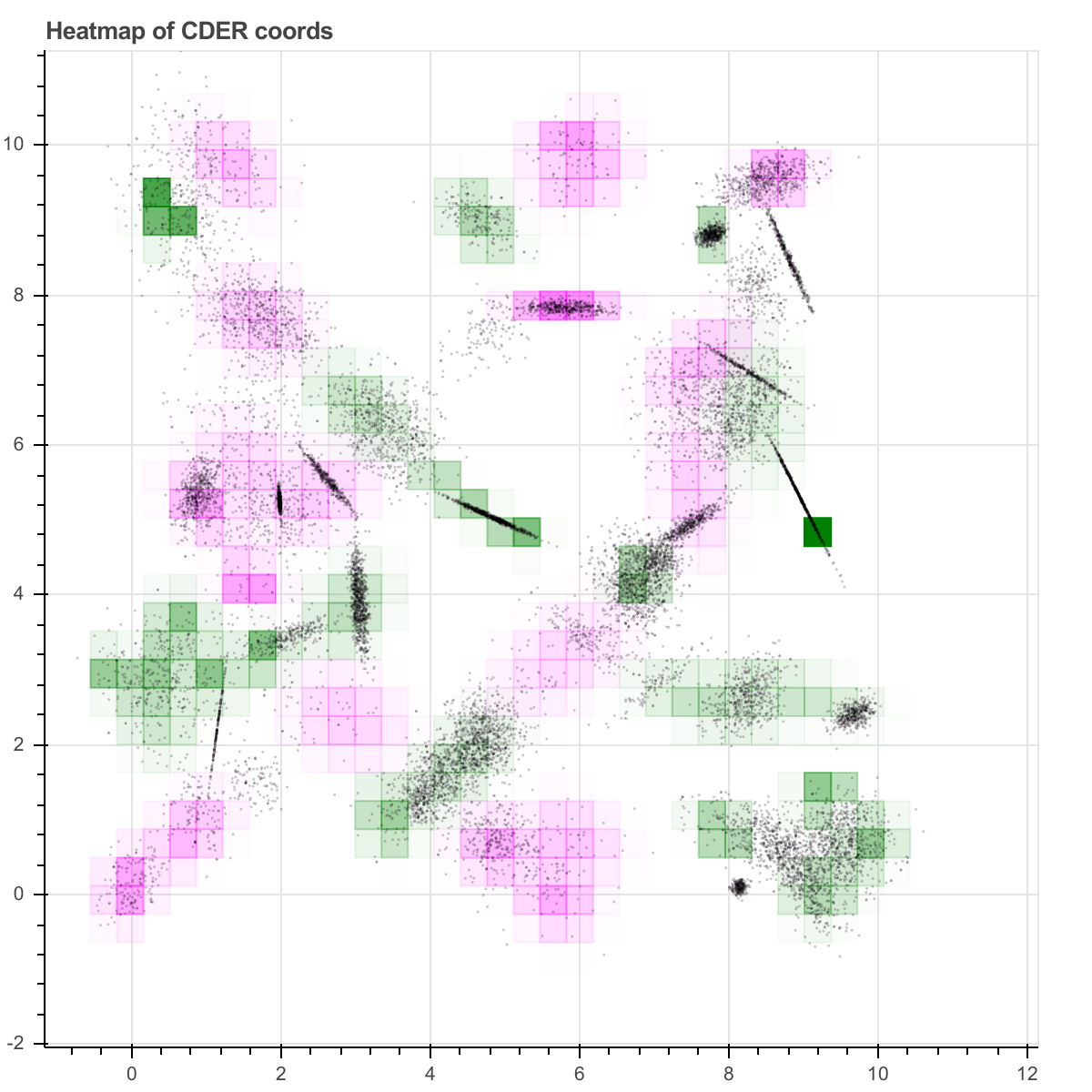}
\end{center}
\caption{The regions and distributional coordinates detected by CDER in the
``deep field'' example.}\label{fig:deepfieldoutput}
\end{figure}

The CDER algorithm builds a Gaussian mixture model focused on regions of low
entropy (Figure~\ref{fig:deepfieldoutput}), so it should not be surprising that
it builds Gaussians near the original Gaussians, as in
Figure~\ref{fig:deepfieldcompare}.

\subsection{Three Labels}
For simplicity, the previous examples have involved only two labels.    In a
two-label system, a low-entropy region has exactly \emph{one} dominant label.
However, the algorithm is sensible for any number of labels, and it may be that
the low-entropy regions are dominated by multiple labels.  Hence, an ensemble
of regions may be necessary to distinguish pointclouds.
\begin{figure}
\begin{center}
\includegraphics[width=0.45\textwidth]{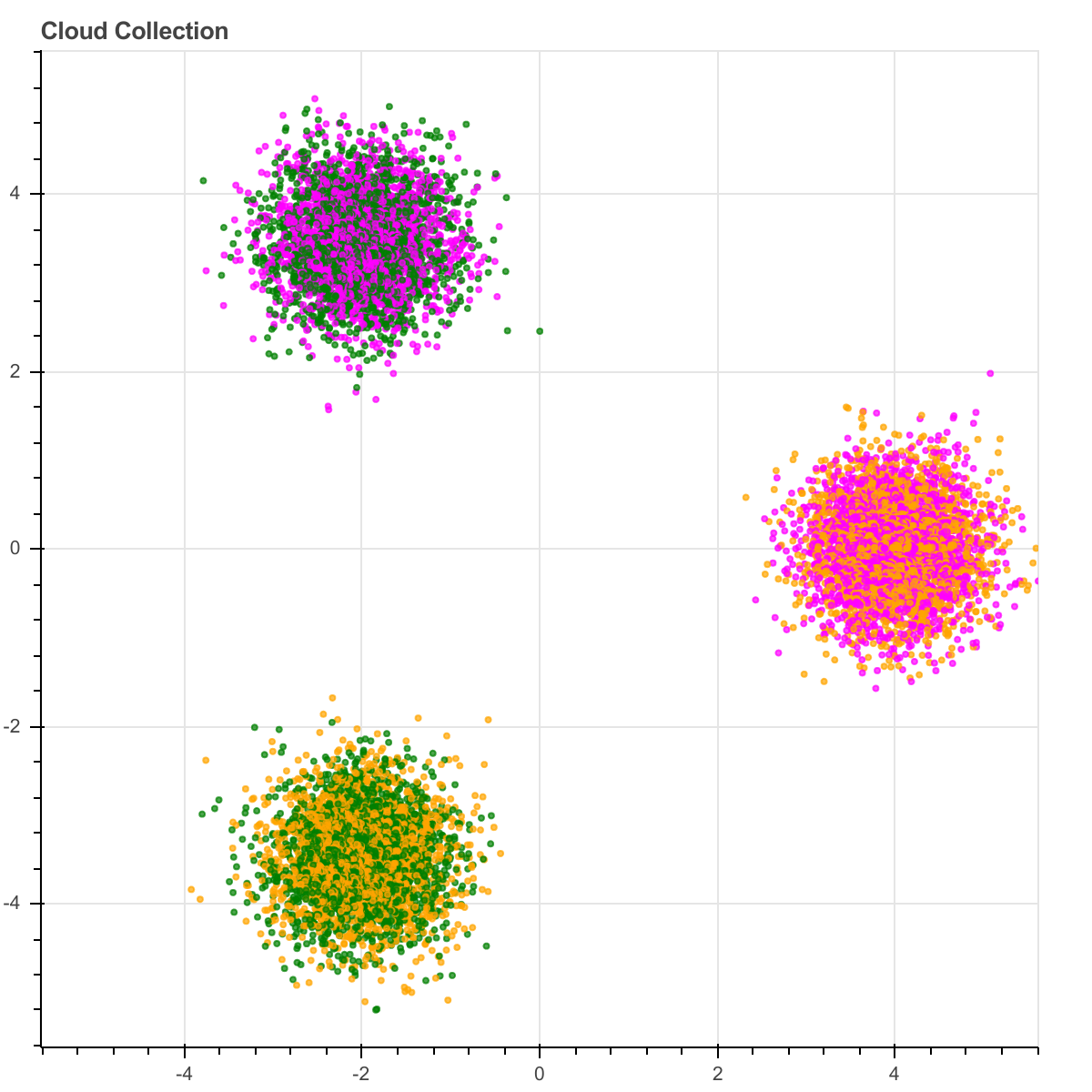}
\includegraphics[width=0.45\textwidth]{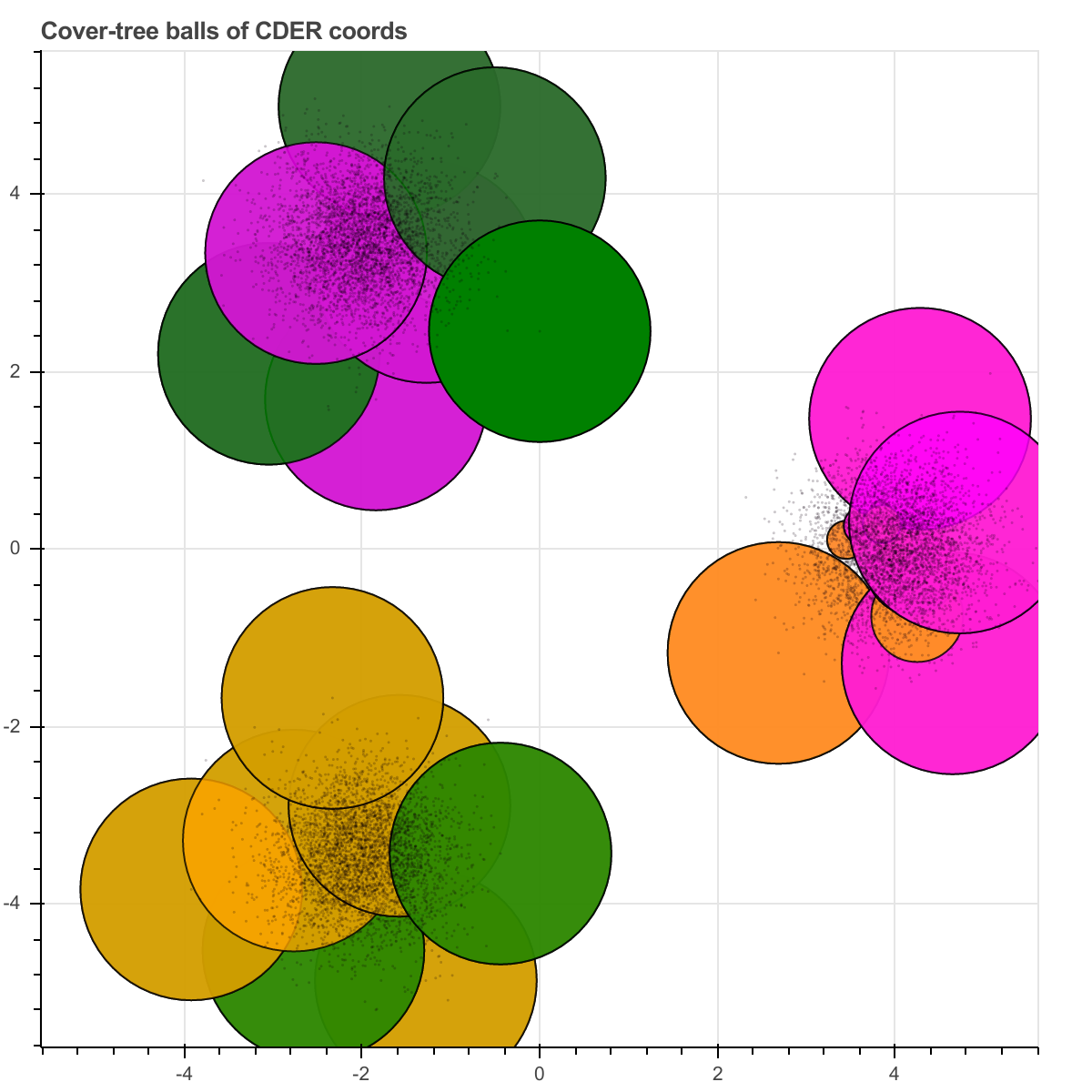}\\
\includegraphics[width=0.45\textwidth]{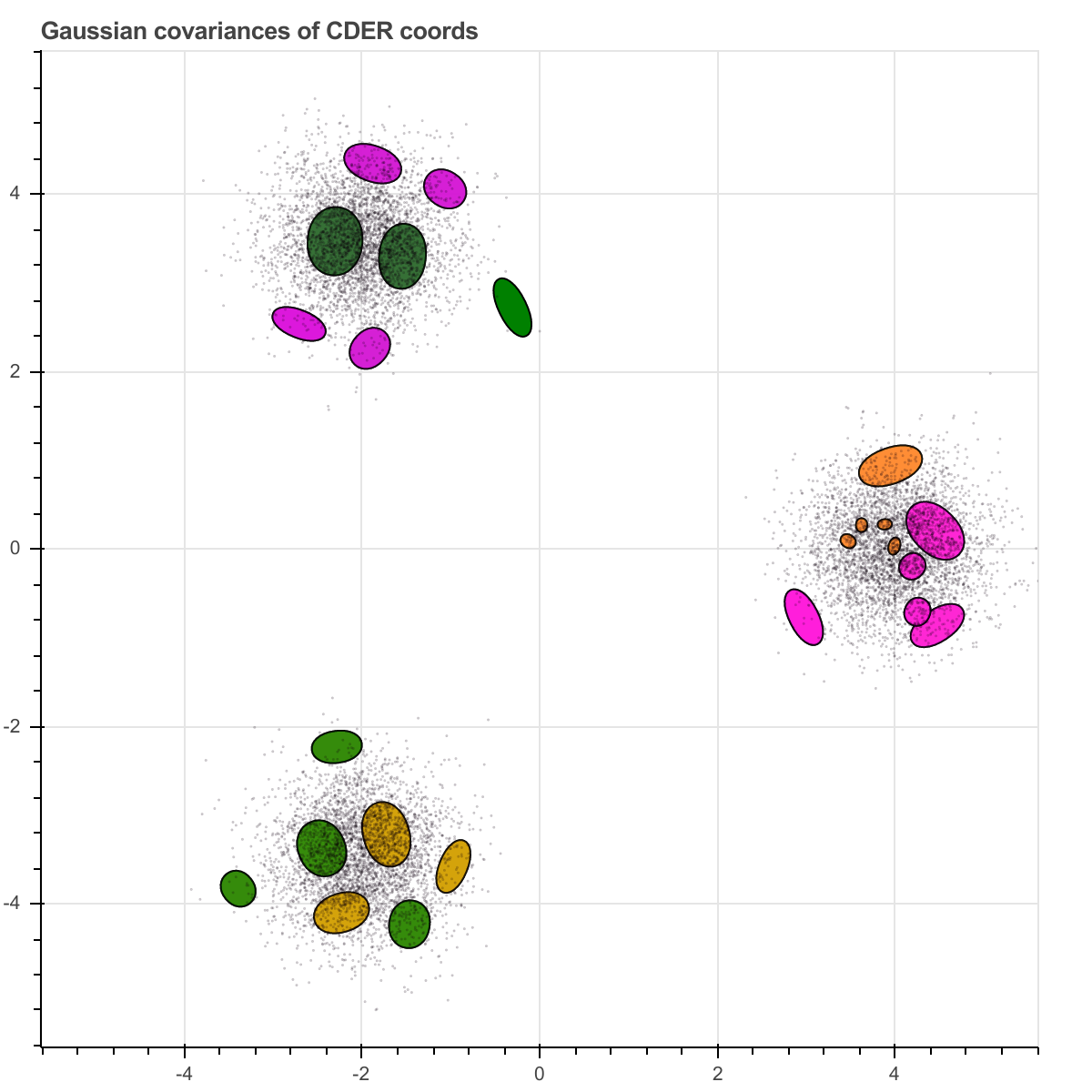}
\includegraphics[width=0.45\textwidth]{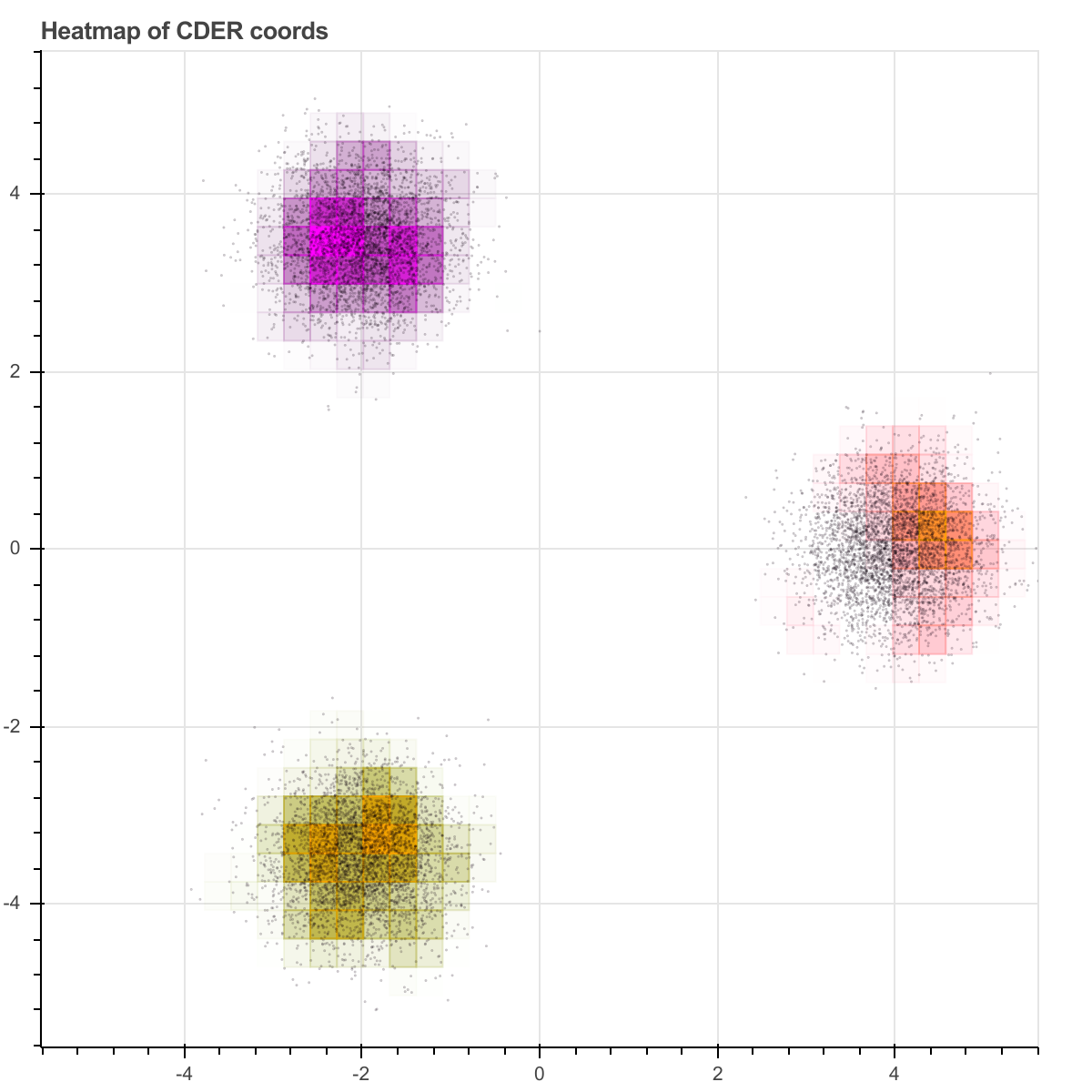}
\end{center}
\caption{A three-label task without unique dominance.
Upper-left: the original cloud collection.  Upper-right: The cover-tree balls
used by CDER.  Lower-left: The Gaussian covariances constructed by CDER.
Lower-right: Heatmap of resulting distributional coordinates.}\label{fig:triple}
\end{figure}

Consider three labels:  0/magenta, 1/green, and 2/orange.
Let $f_0$ be a triple of standard normal distributions: one each at $(0,0)$,
$(4,0)$, and $(-2, 2\sqrt{3})$.
Let $f_1$ be a triple of standard normal distributions: one each at
$(0,0)$,$(-2,2\sqrt{3})$, and $(-2, -2\sqrt{3})$.
$(-2,2\sqrt{3})$, and $(-2, -2\sqrt{3})$.
Let $f_2$ be a triple of standard normal distributions: one each at $(0,0)$,
$(-2,-2\sqrt{3})$, and $(4, 0)$.  See Figure~\ref{fig:triple}.

The CDER algorithm detects these shared regions perfectly, achieving 100\% in a
5-fold cross-validation of this cloud collection with a 80/20 training/testing
split, using the same simple comparison method as in Section~\ref{ex:blobby}.

\section{Stability}\label{sec:stability}
Ideally, one would like to prove that CDER is stable or backward-stable,
in the sense of \cite[Lecture~14]{Trefethen1997}.
To formulate stability or backward-stability, CDER must be expressed as a
numerical approximation of a formal function between normed vector spaces.

Let $C$ denote the vector space of piecewise continuous
functions on
our compact domain.  Define a norm on $C \times C$ by
$\|(f_1,f_2)\| = \sqrt{ \|f_1\|_2^2 + \|f_2\|_2^2}$, where each term is the $L_2$ norm
on our (compact) domain.
Given two ``input ''functions $(f_1, f_2) \in V$ we consider them as
density functions for each of two labels.  Consider the
pointwise entropy,
\[H(x) =
-\frac{f_1(x)}{f_1(x)+f_2(x)}\log_2\left(\frac{f_1(x)}{f_1(x)+f_2(x)}\right)
-\frac{f_2(x)}{f_1(x)+f_2(x)}\log_2\left(\frac{f_2(x)}{f_1(x)+f_2(x)}\right).\]

Let $T$ denote the formal function $T: C
\times C \to C \times C$ by
$T(f_1, f_2) = (g_1, g_2)$
where
\begin{equation}g_1(x)= \begin{cases}
H(x)f_1(x),&\text{if $f_1(x) > f_2(x)$,}\\
0,&\text{otherwise}.
\end{cases}
\end{equation}
and
\begin{equation}g_2(x)= \begin{cases}
H(x)f_2(x),&\text{if $f_2(x) > f_1(x)$,}\\
0,&\text{otherwise}.
\end{cases}
\end{equation}
A numerical algorithm $\tilde{T}$ for $T$ can be obtained roughly as ``sample
equally from $f_1$ and $f_2$, and apply CDER to generate weighted
distributional coordinates.  The weighted sum of the distributional coordinates
for label 1 is $g_1$, and the sum of the distributional coordinates for label 2
is $g_2$.''  A stability result such as this is a subject of future work. We
have not yet succeeded in proving either stability statement for CDER, but the
high cross-validation of the examples above is promising.

\section{Discussion} \label{sec:discussion}

This paper introduced CDER, a data-driven, label-driven feature extraction method for collections of labeled pointclouds. CDER is fast, both in terms of theoretical complexity
and in initial tests on examples. It does not require the user to choose any tuning parameters, and the geometric meaning of its output features are transparently clear.
This section outlines some future directions and makes a few generalizing remarks.

The attentive reader of Section \ref{sec:covertree} may notice that cover trees can be defined in an arbitrary metric space, and indeed this is how they
were originally defined \cite{Har-Peled}.
We use them to construct distributional coordinates, and our algorithm for doing so (Section \ref{sec:cder}) demands that we be able to
quickly compute means of sets of points. While the Fr\'echet Mean (\cite{Burago2001course}, IX) can be defined in any metric space, there are not always fast algorithms for its computation (it also need not exist in all cases).
And so while more generality could be achieved in the definition
of CDER, we only make our complexity statements for cloud collections in a common Euclidean space.

All examples in Section \ref{sec:examples} were artificial, and simply intended to help visualize the novel and fairly-technical CDER
algorithm and emphasize its key properties.
Future work will involve applications of CDER to real data from a variety of real fields, including large administrative datasets of interest to social scientists
and to vehicle tracking \cite{Bendich2016tracking}.

We also hope that CDER will prove useful as a feature-extraction method in topological data analysis \cite{PersistentSurvey},
since persistence diagrams can be thought of as point clouds in the plane. Future work will compare the performance of CDER against other such feature-extraction methods (for example, \cite{reininghaus2015stable} and \cite{Adcock2013}).

Finally, we recall the weighting discussion in Section \ref{sec:weight}, where we used the simplifying assumptions that each color/label was equally likely and that each point
within a single point cloud should be given equal weight. We note that CDER can be easily adapted to accommodate other prior assumptions about relative likelihoods of labels
or even prior assumptions about outlier status of certain points in a cloud, say as part of a Bayesian learning process.

\clearpage

\bibliographystyle{plain}
\bibliography{cder}
\end{document}